\documentclass{article} % For LaTeX2e
\usepackage{iclr2027_conference,times}

\usepackage{amsmath,amsfonts,bm}

\def\eqref#1{equation~\ref{#1}}
\def\1{\bm{1}}

\DeclareMathAlphabet{\mathsfit}{\encodingdefault}{\sfdefault}{m}{sl}
\SetMathAlphabet{\mathsfit}{bold}{\encodingdefault}{\sfdefault}{bx}{n}

\usepackage{hyperref}
\usepackage{url}
\usepackage{booktabs}       % professional-quality tables
\usepackage{amsfonts}       % blackboard math symbols
\usepackage{nicefrac}       % compact symbols for 1/2, etc.
\usepackage{microtype}      % microtypography
\usepackage{xcolor}         % colors
\usepackage{natbib}
\usepackage{amsmath}
\usepackage{enumitem}
\usepackage{tabularx}
\usepackage{graphicx}
\usepackage{multirow}
\usepackage{pifont}
\usepackage{listings}
\usepackage{float}
\usepackage{wrapfig}

\usepackage[most]{tcolorbox}
\tcbuselibrary{listings,breakable,skins}

\definecolor{PromptBg}{HTML}{FAFAFA}
\definecolor{PromptFrame}{HTML}{D9DEE7}
\definecolor{PromptTitleBg}{HTML}{EEF2F7}
\definecolor{PromptKeyword}{HTML}{1F4E79}
\definecolor{PromptComment}{HTML}{5F6B7A}

\lstdefinestyle{promptstyle}{
  basicstyle=\scriptsize\ttfamily,
  breaklines=true,
  breakatwhitespace=false,
  columns=fullflexible,
  keepspaces=true,
  showstringspaces=false,
  frame=none,
  xleftmargin=0pt,
  xrightmargin=0pt,
  aboveskip=0pt,
  belowskip=0pt,
}

\newtcblisting{promptbox}[2][]{
  enhanced,
  breakable,
  colback=PromptBg,
  colframe=PromptFrame,
  coltitle=black,
  colbacktitle=PromptTitleBg,
  fonttitle=\bfseries\small,
  title={#2},
  listing only,
  listing options={
    style=promptstyle,
    language=
  },
  sharp corners=south,
  rounded corners=north,
  boxrule=0.5pt,
  left=1.5mm,
  right=1.5mm,
  top=1mm,
  bottom=1mm,
  #1
}

\title{STRAND: Benchmarking and Improving Object-Centric Spatio-Temporal Monitoring in Video Large Language Models}

\author{Thong Nguyen\thanks{Corresponding author: thong.nguyen@u.nus.edu}\;, \quad Tri Cao, \quad Khoi Le, \quad Cong-Duy Nguyen, \\ {\bf Quynh Vo, \quad See-Kiong Ng, \quad Bryan Hooi Kuen-Yew} \\
National University of Singapore (NUS), Singapore, \\
Centre for AI Research, VinUniversity
}

\iclrfinalcopy % Uncomment for camera-ready version, but NOT for submission.
\begin{document}

\maketitle

\begin{abstract}
While multimodal large language models (MLLMs) have advanced video understanding, they remain highly prone to hallucinations in dynamic scenes. We argue this stems from a failure in spatio-temporal monitoring, the ability to persistently track object identities, states, and relations over time. Existing benchmarks obscure this deficit by relying on single final-answer evaluations for queries that can often be resolved via local visual cues or statistical priors. To rigorously diagnose this, we introduce \textbf{STRAND}, a benchmark of human-verified object-centric facts that evaluates intermediate reasoning by decomposing queries into sub-questions, distinguishing genuine temporal understanding from coincidental correctness. Crucially, we score models with \emph{Faithful Accuracy}, an unconditional joint metric that credits a prediction only when the target answer and every prerequisite sub-question are correct, so that a model cannot inflate its score by being selectively consistent on the small subset of targets it happens to answer correctly. To address failure modes exposed by STRAND, we further propose an object-centric framework that explicitly constructs and reasons over structured object trajectories via chunk-wise state extraction and temporal aggregation. Extensive experiments, including backbone-, frame-, call-, and token-matched comparisons against both end-to-end MLLMs and modular video harnesses, demonstrate that our object-centric framework significantly reduces hallucinated answers and improves spatio-temporal reasoning consistency over state-of-the-art MLLMs. The code, model, and data have been made available at \href{https://nguyentthong.github.io/strand}{nguyentthong.github.io/strand}.
\end{abstract}

\section{Introduction}
Multimodal large language models \citep{cui2024survey, tang2025video, zohar2025apollo} have advanced rapidly in video understanding, with gains in semantic reasoning, instruction following and long-context modelling \citep{wang2026videochat, zhang2024llava, wang2026videochat}. However, they still hallucinate plausible but factually incorrect objects, relations and events \citep{lu2025elv, gao2025exploring}. This matters most in deployment, where a correct answer rarely turns on one salient moment but on a coherent account of what persists and what changes across the video.

Video understanding therefore requires reasoning over a dynamic world in which entities persist, interact and change. Faithful interpretation depends on persistent spatio-temporal monitoring: identifying objects, maintaining their identities, tracking their states, and linking them to temporally ordered events \citep{yuan2025videorefer, li2025sti}. Because each operation supplies context for the next, a confusion between similar objects or a misplaced event propagates into later reasoning while the response stays fluent.

\begin{figure}[t]
    \centering
    \includegraphics[width=0.70\textwidth]{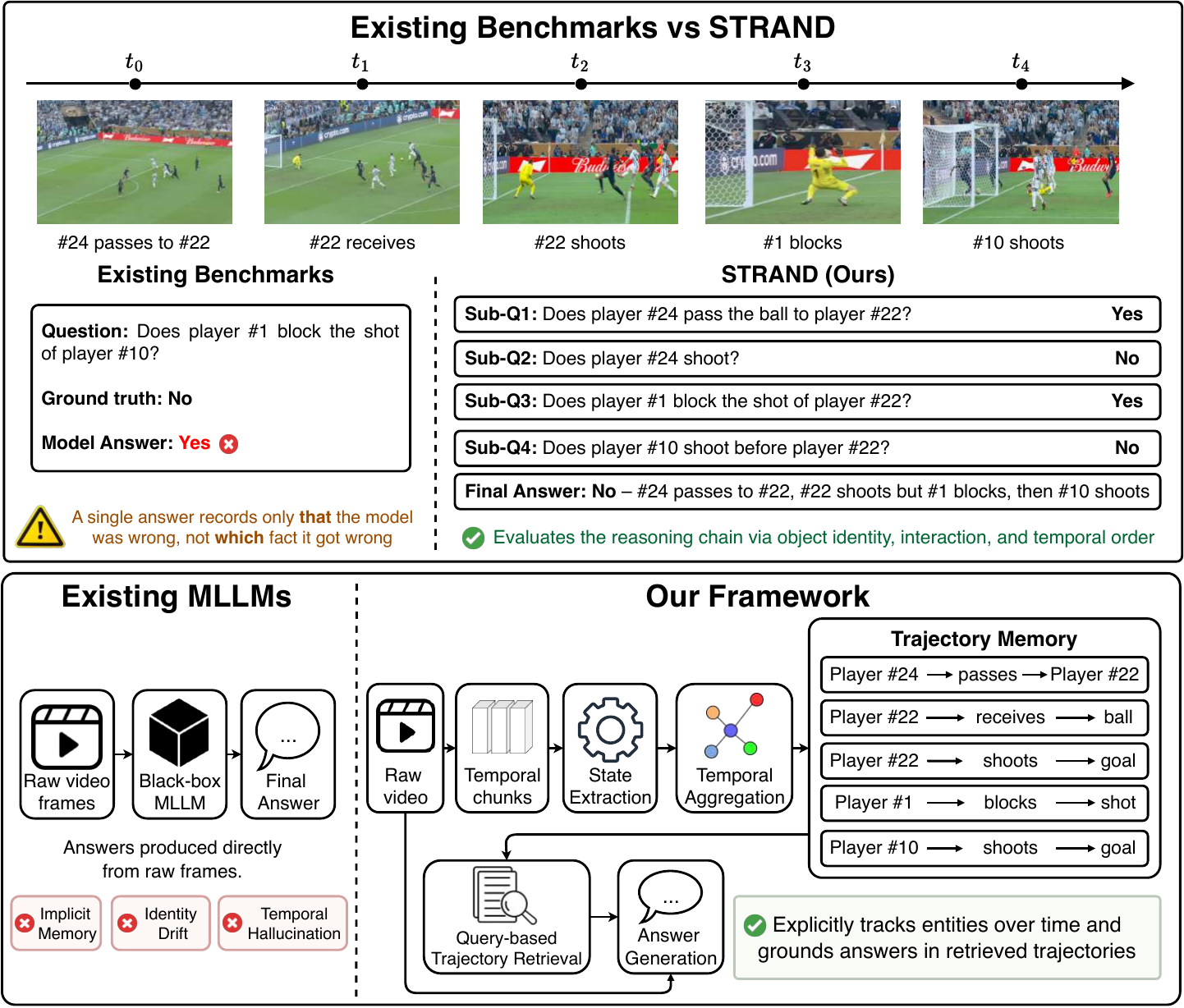}
    \caption{Existing benchmarks and MLLM architectures versus ours. \emph{Top}: a single answer records only that a model is wrong, whereas sub-questions name the facts it rests on. \emph{Bottom}: we answer from object trajectories rather than raw frames alone.}
    \label{fig:teaser}
\end{figure}

This is precisely the ability that current evaluations leave underdetermined. Work on video hallucination \citep{wang2024videohallucer, li2025vidhalluc, rawal2025argus} has sharpened attention to unsupported claims, through fine-grained question-answer pairs over raw frames and subtitles and through mitigation methods that suppress unlicensed statements \citep{xu2025mitigating, bae2025mash, sun2026smartsight}. Other benchmarks instead raise difficulty, widening the temporal horizon \citep{mangalam2023egoschema, wang2025lvbench}, chaining evidence across grounded moments \citep{chen2025grounded}, or making the query object-centric \citep{wang2025object}. These efforts make evaluation more demanding but do not reveal the evidence path by which a model reaches its answer, and standard grading compounds the opacity by checking only the final answer \citep{rawal2025argus, cheng2026v, li2026timeblind}. No amount of added difficulty repairs that criterion, since a model that reaches the right outcome after misbinding an object or reversing two events earns the credit given to one that reasons faithfully. What is missing is auditable evidence rather than harder questions.

We therefore introduce \textbf{STRAND}, which makes the intermediate facts part of the evaluation. Rather than generating question-answer pairs from raw video, it begins from a human-verified world model of each video covering objects, their states, their relations, and the moments these change. Compositional target questions are composed from that structure and paired with atomic sub-questions probing the monitoring steps the target depends on. Because a target and its sub-questions come from the same record, we can ask not only whether a model answers the target but whether it also reports the supporting facts. We mark a target correct only when the model answers it and all of its sub-questions correctly, and faithful accuracy is the fraction of all targets so marked. The denominator includes every target, so a model cannot raise its score by answering only an easy subset.

Because STRAND credits an answer only when the facts supporting it are correct as well, we design an answering system that represents those facts explicitly rather than leaving them implicit. Our framework represents a video as object trajectories, each recording how one object's state changes over time. It divides the video into temporal chunks and extracts object states within each, then links observations across chunks so that one object receives one identity. Given a question, it retrieves the trajectories that question depends on and produces an answer from them. Because this representation is built once per video, the same trajectories serve every question about that video.

Using STRAND, we evaluate current video models and our framework. Frontier models frequently answer a target correctly while getting its supporting facts wrong, so their faithful accuracy falls well below their target accuracy. Our framework narrows this gap, and the gain persists under matched-budget controls that equalise the frame budget, the number of model calls, or the token budget in turn. Answering from trajectories alone costs the stronger backbone little accuracy, so the trajectories already carry what a question needs, whereas permuting their temporal order drops target accuracy from 77.1 to 65.3 on one backbone and from 74.0 to 54.1 on another. The gain therefore rests on the temporal structure of the representation rather than on structured preprocessing alone.

\section{Related Work}
\label{sec:related}

\noindent\textbf{Video Multimodal Large Language Models.} Recent years have seen rapid progress in multimodal large language models (MLLMs), evolving from general vision-language assistants into specialized systems capable of long-context temporal reasoning \citep{cui2024survey, tang2025video}. Early models focused on instruction tuning and short-video understanding \citep{wang2026videochat, zhang2024llava}, while newer architectures emphasize long-context scaling, temporal grounding, and reasoning \citep{li2026videochat, wang2025internvideo2, zohar2025apollo}. This advancement is driven by strong open-source/open-weight models such as InternVL3, Qwen3-VL, Qwen3.5-Omni, and Gemma-4 \citep{zhu2025internvl3, bai2025qwen3, team2026qwen3, team2026gemma}, alongside reasoning-focused video MLLMs including Cosmos-Reason2, VideoRFT, and Video-R1 \citep{azzolini2025cosmos, wang2026videorft, feng2026video}. Proprietary frontier models, including Gemini 3, GPT-5, and Claude-4.6-Sonnet, further continue to push multimodal reasoning capabilities \citep{team2023gemini, achiam2023gpt, liu2026dive}.

\noindent\textbf{Video Hallucination Benchmarks and Mitigation.}
As video MLLMs advance, growing efforts focus on understanding and reducing their failure modes. Existing benchmarks target different vulnerabilities, including object hallucination \citep{wang2024videohallucer}, action and temporal grounding errors \citep{li2025vidhalluc}, structural reasoning deficits \citep{rawal2025argus, cheng2026v}, and temporal drift in long videos \citep{li2026timeblind}. Mitigation approaches span training-time alignment \citep{xu2025mitigating, bae2025mash}, inference-time decoding strategies \citep{sun2026smartsight}, and retrieval-augmented architectures \citep{gao2026mitigating, lu2025elv, gao2025exploring}.

\noindent\textbf{Long-Video and Multi-Hop Reasoning Benchmarks.} A parallel line of work raises task difficulty. EgoSchema \citep{mangalam2023egoschema} and LVBench \citep{wang2025lvbench} widen the temporal horizon, MultiHop-EgoQA \citep{chen2025grounded} chains evidence across grounded moments, and VideoInfer \citep{wang2025object} makes the query object-centric. These raise the bar on what a model must integrate, but they still grade the final answer alone, so a correct response does not show that the model followed the right entities through time. STRAND differs in scoring the supporting facts jointly with the answer.

\noindent\textbf{Object-Centric and Structured Video Understanding.} Recent work explores object-centric and structured video reasoning. VideoRefer \citep{yuan2025videorefer} and STI \citep{li2025sti} show that modelling spatio-temporal object interactions improves grounding, and structured pipelines such as VideoMind \citep{liu2026videomind}, TraveLER \citep{shang2024traveler} and SeViLA \citep{yu2023self} decompose video reasoning into planning, temporal localisation, evidence verification and answer generation.

\section{Object-Centric Spatio-Temporal Formulation}
\label{sec:formulation}
Video understanding is commonly formulated as a direct mapping $A = F_\theta(V, Q)$, where $V = \{f_t\}_{t=1}^{T}$ is a sequence of frames and $Q$ a question. The model compresses the whole video into one latent representation and reasons implicitly within it. This is effective for short clips, but when $V$ spans multiple scenes it entangles object identity, temporal dynamics and reasoning signal, leaving no explicit mechanism to preserve entities or track how their states evolve.

We therefore represent a video as a set of persistent entities $\mathcal{O} = \{o_1, \dots, o_K\}$, where an object is any semantically coherent entity trackable over time. Each $o$ carries a trajectory $\mathcal{S}_o = \{(t, s_t^{(o)}) \mid t \in \mathcal{T}_o\}$ and $\mathcal{S} = \{\mathcal{S}_o\}_{o \in \mathcal{O}}$, with $s_t^{(o)}$ the observable state of $o$ at $t$, covering unary properties such as \texttt{red shirt} and typed relations such as \texttt{passing to $o_2$}. Appendix~\ref{app:extended} states what we register as an object and what a state may contain. Identity is thus separated from state, and video answering becomes structured reasoning over trajectories, $A = \Psi_Q(\mathcal{S})$. This is the abstraction both the benchmark and the framework are built on.

\section{STRAND Construction}

\begin{figure}[t]
    \centering
    \includegraphics[width=0.80\textwidth]{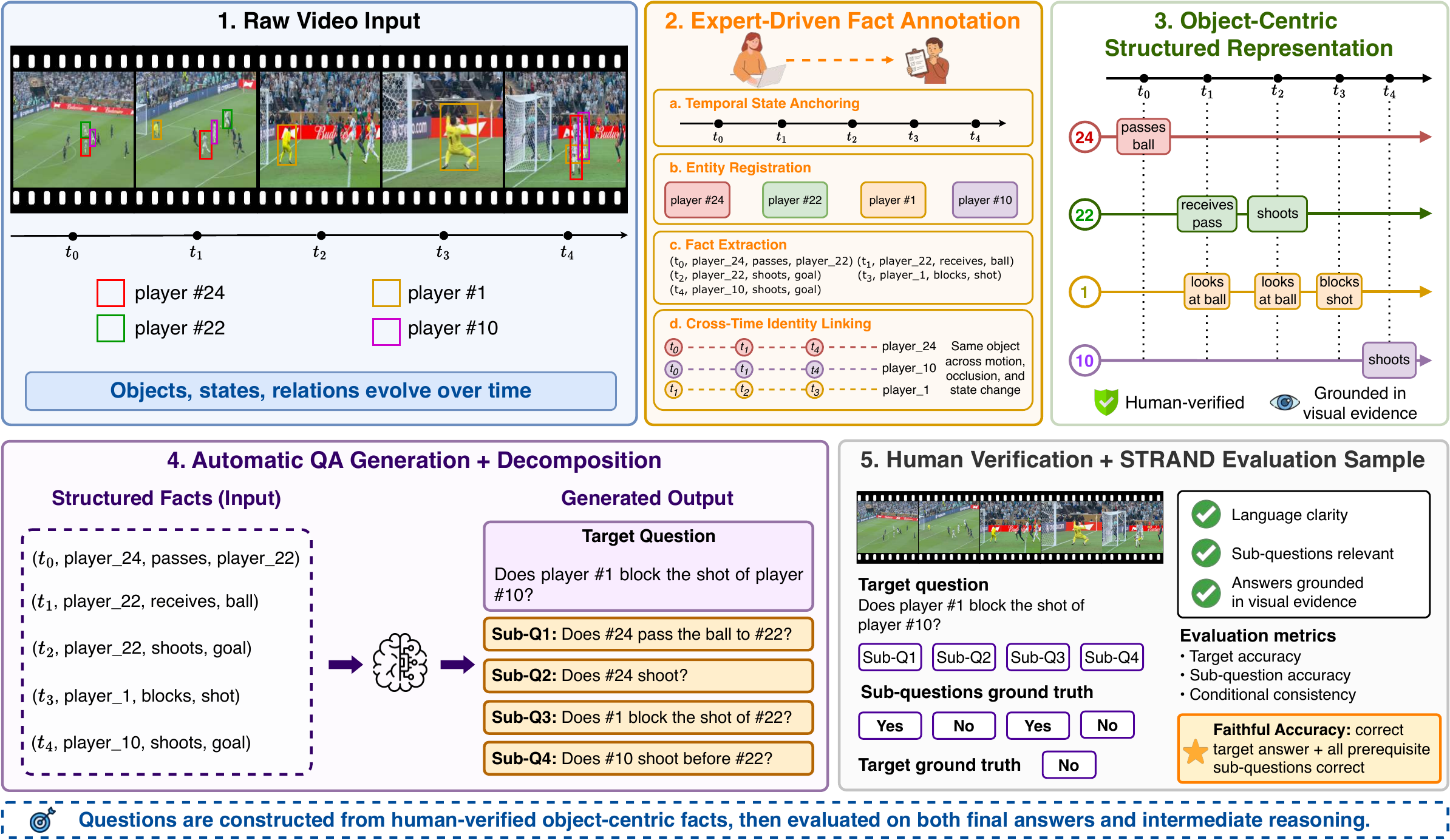}
    \caption{Overview of the STRAND construction pipeline. Annotators anchor timestamps, register entities, extract typed facts, and link identities across time. Target questions and sub-questions are composed from that record and verified.}
    \label{fig:benchmark}
\end{figure}

From the object-centric perspective, faithful video understanding requires an internal monitor over objects, states, interactions and identities across time, which existing datasets do not evaluate. Many question-answer pairs are solvable from one frame or from language priors, and grading the final answer alone hides whether a model resolved the underlying temporal dependencies or guessed.

\subsection{Expert-Driven Fact Annotation}
To prevent the benchmark from inheriting hallucinatory details from automated generators, we decouple the extraction of the world state from the generation of the QA pairs. We use a closed pool of 50 trained annotators to construct explicit factual representations of the videos. Annotators train extensively on the required schema, and inter-annotator agreement reaches 0.88 (Cohen's kappa), indicating a strong consensus on object identities and state transitions. Annotators follow a strict protocol to isolate directly observable visual evidence from subjective inferences:

Annotation proceeds in four steps. Annotators anchor a discrete set of key timestamps $\mathcal{T}$ at salient interactions, state transitions and occlusions. They register the task-relevant entities $\mathcal{O}$ with canonical, visually grounded identifiers such as \texttt{player\_24}, record observable states and relations at each timestamp as typed tuples such as $(t, o_1, \texttt{passing\_to}, o_2)$, and trace entities across the timeline to bind identities into trajectories $\mathcal{S}_o$. This yields a human-verified structured record of the video's dynamics.

\subsection{Automatic QA Generation and Decomposition}
\label{sec:decomposition}
Given the annotated facts, we use a Large Language Model (LLM) to organize them into the global object-centric representation $\mathcal{S}$ and generate reasoning chains. The generator operates only on the symbolic fact tuples and never sees the video, so errors it introduces are ones of phrasing or unsupported inference rather than invented visual detail. Its role is limited to chain selection and natural-language surface realization. We use Llama-3.1-70B-Instruct, which is disjoint from every model family we evaluate, so no evaluated system shares pretraining lineage with the model that phrases its questions.

\noindent\textbf{Target question generation.} Rather than naively combining facts, the LLM constructs reasoning chains over the trajectories. Consider an annotated chain from the video in Figure~\ref{fig:teaser}, a broadcast football sequence: $f_1 = (t_0, \texttt{player\_24}, \texttt{passes}, \texttt{player\_22})$, $f_2 = (t_1, \texttt{player\_22}, \texttt{receives}, \texttt{ball})$, $f_3 = (t_2, \texttt{player\_22}, \texttt{shoots}, \texttt{goal})$, $f_4 = (t_3, \texttt{player\_1}, \texttt{blocks}, \texttt{shot})$ and $f_5 = (t_4, \texttt{player\_10}, \texttt{shoots}, \texttt{goal})$.

The target question is \emph{``Does player \#1 block the shot of player \#10?''}, with ground truth \emph{No}. Two players shoot and player \#1 blocks once, so the question asks which shot the block stops. Both attributions describe ordinary play and football knowledge favors neither, so only the position of the block distinguishes them: it falls at $t_3$, after \#22's shot at $t_2$ and before \#10 shoots at $t_4$.

\noindent\textbf{Sub-question decomposition.} The LLM decomposes the target into atomic sub-questions probing the facts it draws on: \emph{Does player \#24 pass the ball to player \#22?} (\emph{Yes}), \emph{Does player \#24 shoot?} (\emph{No}), \emph{Does player \#1 block the shot of player \#22?} (\emph{Yes}), and \emph{Does player \#10 shoot before player \#22?} (\emph{No}). The third checks the attribution of the block and the fourth the order of the two shooters. Faithful accuracy requires all four, which is stricter than the minimal set the target logically needs. The target is temporal by construction: two plausible shooters, one block, and no domain preference between the attributions leave only the observed order to identify the shot that is stopped. We compose a target under exactly this condition, whenever its answer turns on the order of two events and both orders stay consistent with the events themselves. Of the 382 temporally constrained targets, 361 ($94.5\%$) carry an ordering sub-question meeting it.

\subsection{Evaluation Metrics}
\label{sec:eval_metrics}

Because STRAND evaluates the final prediction and the internal monitoring of facts, we use a multi-tiered scheme. For a target $q$ with ground truth $y$ and prediction $\hat{y}$, let $S_q = \{s_1, \dots, s_m\}$ be its supporting sub-questions with answers $y^{(j)}_{\text{sub}}$ and predictions $\hat{y}^{(j)}_{\text{sub}}$.

\begin{itemize}[leftmargin=*,noitemsep,topsep=0pt]
    \item \textbf{Target Accuracy ($A_{\text{target}}$):} Standard performance on the complex target questions:
    \begin{equation}
    A_{\text{target}} = \frac{1}{N} \sum_{i=1}^{N} \mathbb{I}(\hat{y}_i = y_i)
    \end{equation}

    \item \textbf{Sub-question Accuracy ($A_{\text{sub}}$):} Measures the model's foundational perception ability by computing accuracy across all sub-questions:
    \begin{equation}
    A_{\text{sub}} =
    \frac{
    \sum_{i=1}^{N}\sum_{j=1}^{m_i}
    \mathbb{I}(\hat{y}^{(j)}_{\text{sub},i}=y^{(j)}_{\text{sub},i})
    }{
    \sum_{i=1}^{N} m_i
    }
    \end{equation}

    \item \textbf{Faithful Accuracy ($A_{\text{faith}}$), primary metric:} Credits a question only when the target answer \emph{and every} supporting sub-question are correct. It is computed over all $N$ targets, without conditioning on the model's own correctness:
    \begin{equation}
    A_{\text{faith}}
    =
    \frac{1}{N}
    \sum_{i=1}^{N}
    \mathbb{I}(\hat{y}_i = y_i)
    \prod_{j=1}^{m_i}
    \mathbb{I}(\hat{y}^{(j)}_{\text{sub},i}=y^{(j)}_{\text{sub},i})
    \end{equation}

    \item \textbf{Conditional Consistency ($A_{\text{cons}}$), secondary:} Evaluates whether the model maintains coherent intermediate reasoning when the final target prediction is correct. For each correctly answered target question, consistency is defined as the proportion of correctly answered sub-questions:
    \begin{equation}
    A_{\text{cons}}
    =
    \frac{1}{|\mathcal{C}|}
    \sum_{i \in \mathcal{C}}
    \frac{1}{m_i}
    \sum_{j=1}^{m_i}
    \mathbb{I}(\hat{y}^{(j)}_{\text{sub},i}=y^{(j)}_{\text{sub},i})
    \end{equation}
    where $\mathcal{C} = \{i \mid \hat{y}_i = y_i\}$ is the set of target questions answered correctly by the model.
\end{itemize}

\noindent\textbf{Why $A_{\text{faith}}$ and not $A_{\text{cons}}$.} $A_{\text{cons}}$ averages over $\mathcal{C}$, whose membership depends on the model, so two models are scored on different subsets. It is maximised by low-recall selectivity and does not fall when a model misses a target outright. $A_{\text{faith}}$ has a fixed denominator, is monotone in both target and sub-question correctness, and is zero for a model that answers nothing, so it admits direct cross-model comparison. We report $A_{\text{cons}}$ alongside it but state all consistency claims in terms of $A_{\text{faith}}$.

A model with high $A_{\text{target}}$ but low $A_{\text{faith}}$ is likely relying on spurious correlations rather than maintaining a robust spatio-temporal monitor.

\subsection{Statistics and chance baselines}
\label{sec:answer_format}

STRAND contains 88 videos averaging 183 seconds, with 977 targets and 2516 sub-questions, or 2.58 per target and 11.1 targets per video. Fifty trained annotators produce the records, with inter-annotator agreement of $0.88$ (Cohen's $\kappa$). All items are yes/no, chosen so that sub-questions stay atomic and grading needs no judge.

Accuracies should be read against blind references rather than against $50\%$. A majority-class predictor reaches $A_{\text{faith}} = 19.1\%$, a question-only predictor given no video reaches $21.4\%$, and a caption-only predictor reaches $31.2\%$. Restricted to the ordering sub-questions, the question-only predictor falls to $49.3\%$, no better than guessing, so the order of events has to be read off the video. Appendix~\ref{app:extended} reports the full breakdown.

\subsection{Human Verification}
\label{sec:human_verification}

Even with structured inputs, LLMs can introduce phrasing ambiguities or hallucinated priors. To ensure benchmark quality, annotators review the generated QA pairs alongside the source video. They verify that (1) the language is clear and natural, (2) the sub-questions are jointly
sufficient to answer the target query and each probes a fact the target draws on, and (3) all answers are unambiguously grounded in visual evidence. Any sample featuring unsupported assumptions or unreliable identity bindings is discarded.

Annotators record whether an alternative sufficient fact set exists and find one for $5.8\%$ of targets, and Appendix~\ref{app:extended} reports the elicitation study behind this.
\section{Object-Centric Trajectory Reasoning}

\begin{figure}[t]
    \centering
    \includegraphics[width=0.90\textwidth]{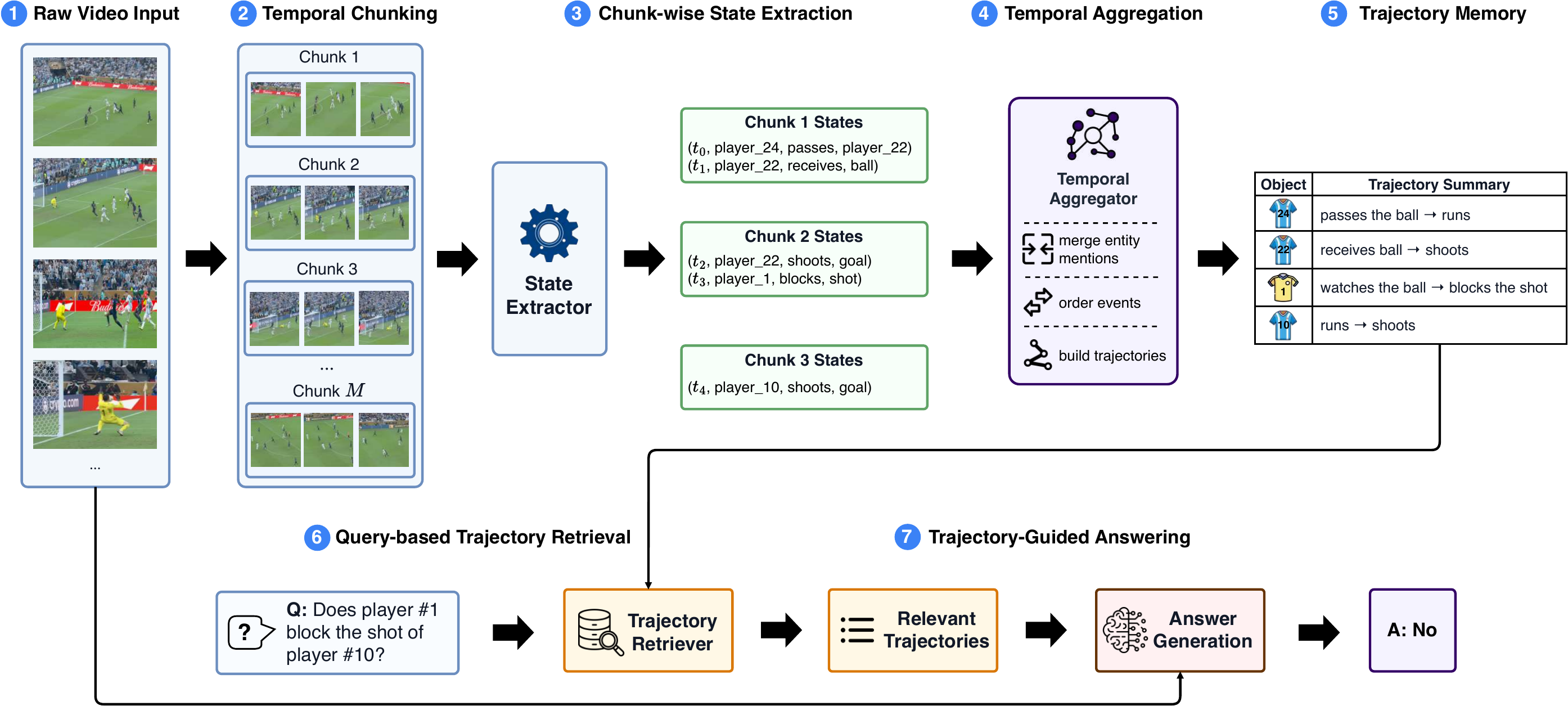}
    \caption{\textbf{Overview of our framework.} Stages 1 to 5 run once per video. Stages 6 and 7 run once per question.}
    \label{fig:strand_method}
\end{figure}

\vspace{-2mm}

Building on the object-centric formulation, we convert a video into structured object trajectories and answer over that representation rather than re-reading the full frame sequence for every query. Given \(V = \{f_t\}_{t=1}^{T}\), we build trajectories $V \xrightarrow{E_\phi} \tilde{\mathcal{S}} \xrightarrow{\mathcal{A}_\psi} \hat{\mathcal{S}} = \{\mathcal{S}_o\}_{o \in \mathcal{O}}$ once per video and reuse them across every question about it, then retrieve and answer per question:
\begin{equation}
(\hat{\mathcal{S}}, Q) \xrightarrow{R_\omega} \mathcal{S}_Q, \qquad A = G_\eta(\mathcal{S}_Q, Q, \bar{V}),
\end{equation}
\noindent where $\bar{V}$ is a fixed uniformly sampled frame budget.

\vspace{-2mm}

\subsection{Chunk-wise state extraction}

We partition the video into \(M\) disjoint chunks $V = \bigcup_{m=1}^{M} V^{(m)}$, sample $N_l$ frames from each, and apply a VLM-based extractor $E_\phi$ to every chunk independently, so extraction parallelises. Writing $\tilde{\mathcal{I}}_m^{(i)}$ for the sampled timestamps at which object $o_i^{(m)}$ is visible,
\begin{equation}
\tilde{\mathcal{S}}^{(m)} = E_\phi(\tilde{V}^{(m)}) = \Big\{\big(o_i^{(m)}, \{(t, s_t^{(o_i^{(m)})})\}_{t \in \tilde{\mathcal{I}}_m^{(i)}}\big)\Big\}_{i=1}^{K_m},
\qquad
\tilde{\mathcal{S}} = \bigcup_{m=1}^{M} \tilde{\mathcal{S}}^{(m)},
\end{equation}
with $K_m$ the number of objects detected in chunk $m$. A state is relational rather than purely attributive, so $s_t^{(o)}$ holds the predicate the object participates in together with its arguments, which is what the tuples of Figure~\ref{fig:strand_method} report.

\vspace{-1mm}

\subsection{Temporal aggregation}
\label{sec:temporal_aggregation}

Chunk-level observations contain duplicate objects, noisy states and identities fragmented across time. A deterministic, symbolic aggregator consolidates them, $\hat{\mathcal{S}} = \mathcal{A}_\psi(\tilde{\mathcal{S}})$ with $\mathcal{S}_o = \{(t, s_t^{(o)}) \mid t \in \mathcal{T}_o\}$ ordered in time. The subscript $\psi = (\Delta t_{max}, \tau_{conf})$ collects fixed hyperparameters, not learned weights, and $\mathcal{A}_\psi$ is \emph{not} a VLM in our default configuration: it issues no model call, whereas the LLM-summarize row of Table~\ref{tab:ablations} is an ablation. Its four steps are similarity scoring over visual attributes and spatial proximity, constraint filtering on temporal proximity and a confidence threshold, bipartite conflict resolution, and trajectory generation in which unmatched observations survive as singletons. Appendix~\ref{app:extended} states each step in full, with the two instantiations of $\mathcal{A}_\psi$ we evaluate.

\vspace{-1mm}

\subsection{Retrieval and answering}

Not every trajectory bears on a given question, so we retrieve a query-relevant subset $\mathcal{S}_Q = R_\omega(\hat{\mathcal{S}}, Q) \subseteq \hat{\mathcal{S}}$ and generate the answer from it. Operating over structured trajectories rather than raw frames lets this stage address object identities, state transitions and temporal relations directly. $\bar{V}$ is a fixed uniformly sampled frame budget, independent of the question, so among the video-derived inputs only $\mathcal{S}_Q$ varies per query. For the stronger backbone retrieval acts mainly as a context-length control rather than a source of accuracy, though the weaker one is markedly more sensitive to it, as Appendix~\ref{app:retrieval_analysis} quantifies.

\section{Experiments}
\label{sec:experiments}

\begin{table}[t]
\centering
\caption{STRAND results (\%). \textbf{Best} and \underline{second-best} per column. Four further open-weight models are in Appendix~\ref{app:extended}. $A_{\text{faith}}$ is the primary metric and $A_{\text{cons}}$ is reported for reference only (Section~\ref{sec:eval_metrics}).}
\label{tab:strand_results}
\footnotesize
\begin{tabular}{@{}l|c|ccc@{}}
\toprule
\multirow{2}{*}{\textbf{Method}} & \multicolumn{4}{c}{\textbf{STRAND}} \\
\cmidrule(lr){2-5}
 & $A_{\text{faith}}$ & $A_{\text{target}}$ & $A_{\text{sub}}$ & $A_{\text{cons}}$ \\
\midrule
\multicolumn{5}{@{}l}{\textit{Ours}} \\
Ours (Gemini-3-Flash) & \textbf{59.4} & \textbf{77.1} & \textbf{76.3} & \underline{81.3} \\
Ours (Qwen3-VL-235B) & \underline{54.2} & \underline{74.0} & 72.6 & 78.1 \\
\midrule
\multicolumn{5}{@{}l}{\textit{Proprietary end-to-end MLLMs}} \\
Gemini-3.1-Pro & 38.9 & 61.4 & 64.9 & 69.2 \\
Gemini-3-Flash & 28.7 & 48.7 & 61.7 & 69.9 \\
Gemma-4-27B & 35.8 & 55.5 & 60.0 & 72.7 \\
GPT-5 & 20.0 & 32.5 & 43.2 & 70.5 \\
Claude-4.6-Sonnet & 30.2 & 48.3 & 62.9 & 73.2 \\
\midrule
\multicolumn{5}{@{}l}{\textit{Open-weight end-to-end MLLMs}} \\
InternVL3-78B & 29.7 & 49.9 & 66.1 & 71.0 \\
Qwen3-VL-32B Think & 46.1 & 65.0 & 70.6 & 78.1 \\
Qwen3.5-27B & 45.5 & 68.4 & 66.7 & 73.3 \\
Cosmos-Reason2-8B & 41.5 & 54.7 & \underline{74.7} & \textbf{83.7} \\
VideoRFT-7B & 16.4 & 38.6 & 55.1 & 50.8 \\
Video-R1-7B & 4.3 & 24.8 & 19.6 & 21.9 \\
\midrule
\multicolumn{5}{@{}l}{\textit{Structured pipelines}} \\
VideoMind-7B & 18.9 & 50.3 & 37.9 & 43.2 \\
TraveLER & 22.0 & 50.6 & 41.8& 49.3\\
SeViLA & 26.1 & 51.4 & 59.6 & 56.8 \\
\bottomrule
\end{tabular}
\vspace{-10pt}
\end{table}

\subsection{Experimental Setup}
\label{sec:exp_setup}

\textbf{Benchmarks and metrics.}
We evaluate our framework on four benchmarks: STRAND, VideoHallucer \citep{wang2024videohallucer}, Video-MME \citep{fu2025video}, and EgoSchema \citep{mangalam2023egoschema}. In the main paper, we focus on STRAND using the metrics defined in Section~\ref{sec:eval_metrics}, namely Faithful Accuracy ($A_{\text{faith}}$, primary), Target Accuracy ($A_{\text{target}}$), Sub-question Accuracy ($A_{\text{sub}}$), and conditional Consistency ($A_{\text{cons}}$). Results on the remaining benchmarks are provided in Appendix~\ref{app:other_benchmark}.

\textbf{Implementations.}
We instantiate our framework in two configurations. \textbf{Ours (Gemini-3-Flash)} uses \texttt{gemini-3-flash-preview} for $E_\phi$, $R_\omega$ and $G_\eta$. \textbf{Ours (Qwen3-VL-235B)} uses \texttt{qwen3-vl-235b-a22b-thinking} for $E_\phi$, \texttt{qwen3-235b-a22b} for $R_\omega$ and \texttt{qwen3.5-27b} for $G_\eta$. Unless ablated, both use 15-second chunks, 60 frames per chunk, 64 frames for $G_\eta$, and identity linking. All prompts appear verbatim in Appendix~\ref{app:prompts}, for our framework and every baseline.

\subsection{Main Results}
\label{sec:main_results}

Table~\ref{tab:strand_results} shows that STRAND is challenging for frontier MLLMs. No end-to-end model exceeds $68.4\%$ $A_{\text{target}}$, several fall below $50\%$, and single-frame performance drops to near-random for the models audited in Appendix~\ref{app:shortcut_audit}. VideoRFT-7B, Video-R1-7B and VideoMind-7B fall below the majority-class $A_{\text{faith}}$ of $19.1\%$ (Section~\ref{sec:answer_format}), so without reliable temporal tracking they cannot ground a target answer in the facts it rests on. The Gemini-3-Flash instantiation reaches $59.4\%$ $A_{\text{faith}}$, $77.1\%$ $A_{\text{target}}$, $76.3\%$ $A_{\text{sub}}$ and $81.3\%$ $A_{\text{cons}}$, ahead of Gemini-3.1-Pro by $20.5$, $15.7$, $11.4$ and $12.1$ points. That margin sets the scale of the effect rather than isolating its cause, which the ablations below and Appendix~\ref{app:budget_matched} address.

\noindent\textbf{Temporal ordering remains the bottleneck.} Splitting $A_{\text{sub}}$ by probe type locates the residual errors. The Gemini-3-Flash instantiation reaches $78.8\%$ on sub-questions about individual facts but $61.4\%$ on those about the order of two events, and with Qwen3-VL-235B it reaches $75.4\%$ against $56.2\%$. The gap of roughly $18$ points holds across both backbones, so even a system that builds explicit trajectories identifies events far more reliably than it orders them. Ordering is the capability STRAND isolates and the one that remains open.

Cosmos-Reason2-8B has the highest $A_{\text{cons}}$ while answering $22.4$ points fewer targets correctly, which is why we state all consistency claims in terms of $A_{\text{faith}}$ (Appendix~\ref{app:extended}).

Significance is assessed by paired bootstrap over STRAND target accuracy (10{,}000 resamples, clustered at the video level). Each instantiation improves significantly over the strongest end-to-end model of its own family ($p < 0.005$). Against the strongest open-weight model by target accuracy, Qwen3.5-27B, the Gemini instantiation gains $[+6.8\%, +10.6\%]$ at $95\%$ confidence.

\noindent\textbf{Transfer to external benchmarks.} Table~\ref{tab:external} condenses the external evaluation, with full sub-scores in Appendix~\ref{app:other_benchmark}. The Gemini-3-Flash instantiation leads the strongest baseline on VideoHallucer by $9.3$ points and on EgoSchema by $1.6$, and trails Gemini-3.1-Pro on Video-MME by $0.5$. The Qwen instantiation improves on its extractor backbone run end-to-end by $11.6$ points on VideoHallucer. Since STRAND is built around object-centric trajectories, these results check generality rather than claim state of the art.

\begin{table}[t]
\centering
\caption{Condensed external-benchmark results (\%). Full sub-scores in Appendix~\ref{app:other_benchmark}.}
\label{tab:external}
\footnotesize
\begin{tabular}{@{}l|ccc@{}}
\toprule
\textbf{Method} & \textbf{VideoHallucer} & \textbf{Video-MME} & \textbf{EgoSchema} \\
\midrule
Ours (Gemini-3-Flash) & \textbf{76.8} & \underline{76.8} & \textbf{78.4} \\
Ours (Qwen3-VL-235B)  & \underline{71.4} & 73.6 & 76.4 \\
\midrule
Gemini-3.1-Pro & 63.1 & \textbf{77.3} & \underline{76.8} \\
GPT-5 & 67.5 & 76.2 & 75.4 \\
Qwen3-VL-235B & 59.8 & 72.6 & 72.1 \\
LLaVA-OV & 48.5 & 66.3 & 60.1 \\
\bottomrule
\end{tabular}
\end{table}

\noindent\textbf{Efficiency.} Our framework trades preprocessing and per-query compute for accuracy. Trajectory construction is \emph{amortised}: the $M$ extraction calls are paid once per video and reused across every query on it. Because $G_\eta$ consumes $64$ frames as well as the retrieved state, our per-query calls are slightly heavier than an end-to-end call. With $Q = 39.7$ questions per video, counting targets and sub-questions, Appendix~\ref{app:efficiency} reports latency and token cost split into one-off and per-query components.

\subsection{Ablation Study}
\label{sec:ablations}

\begin{table}[t]
\centering
\caption{Ablation studies on STRAND Target Accuracy (\%).}
\label{tab:ablations}
\setlength{\tabcolsep}{3pt}
\resizebox{\textwidth}{!}{%
\begin{tabular}{@{}llcccc@{}}
\toprule
\textbf{Ablation Focus} & \textbf{Pipeline} & \textbf{Chunk (s)} & \textbf{Frames/chunk} & \textbf{Frames@$G_\eta$} & \textbf{$A_{\text{target}}$ (\%)} \\
\midrule
\multicolumn{6}{l}{\textit{Reference Base Config}} \\
\quad Base Configuration & Ours (Gemini-3-Flash) & 15 & 60 & 64 & 77.1 \\
\quad Base Configuration & Ours (Qwen3-VL-235B)  & 15 & 60 & 64 & 74.0 \\
\midrule
\multicolumn{6}{l}{\textit{Query-based retrieval}} \\
\quad No filter (full timeline) & Gemini / Qwen & 15 & 60 & 64 & 77.0 / 64.6 \\
\quad Random filter (top-$k$)   & Gemini / Qwen & 15 & 60 & 64 & 76.5 / 68.0 \\
\midrule
\multicolumn{6}{l}{\textit{Temporal aggregation}} \\
\quad Aggregator: LLM-summarize & Gemini / Qwen & 15 & 60 & 64 & 76.7 / 62.7 \\
\bottomrule
\end{tabular}
}
\vspace{-10pt}
\end{table}

\noindent\textbf{Matched-budget comparison.} A harness brings extra compute as well as a different representation, so we hold the backbone fixed and equalise one resource at a time. In $A_{\text{faith}}$, our framework reaches $54.2$ against $41.5$ at an equal 64-frame budget, $49.3$ against $39.8$ at equal model calls, and $52.8$ against $45.1$ at an equal token budget (Appendix~\ref{app:budget_matched}). This, rather than the end-to-end rows of Table~\ref{tab:strand_results}, is what the design claim rests on.

We ablate key components of our framework. Appendix~\ref{app:full_ablation_studies} reports the full set, including the visual-budget and representation-form ablations summarised below.

\noindent\textbf{Visual budget} (Appendix~\ref{app:full_ablation_studies}). Lowering the answerer to 8 frames barely moves Gemini-3-Flash but costs Qwen $10.4$ points, and lowering extractor density to 8 frames per chunk degrades both.

\noindent\textbf{Retrieval and aggregation.} Skipping retrieval filtering costs Qwen $9.4$ points and Gemini $0.1$, so only the weaker backbone is sensitive to timeline clutter. \textbf{Aggregation Method:} Symbolic identity linking outperforms LLM summarisation by never re-generating a state, so low-salience details survive.

\noindent\textbf{The mechanism is backbone-dependent.} For Qwen3-VL-235B every ablated component is load-bearing, costing $10.4$, $9.4$ and $11.3$ points. For Gemini-3-Flash the same interventions cost $0.6$, $0.1$ and $0.4$, all within a tie, so a simpler explanation survives for the stronger backbone: structured preprocessing with a capable model may account for most of the gain. Two results bear on it. Text-only answering still reaches $73.6$ for Gemini, and shuffling the temporal order of the same representation costs $11.8$ points ($77.1 \rightarrow 65.3$). The shuffle separates temporal order from unordered structure, but not object-centric structure from generic structure. We therefore restrict the strong per-stage claim to Qwen. Appendix~\ref{app:extended} names the controls we lack.

\subsection{Qualitative Analysis}
\label{sec:qualitative}

The paired evaluation exposes shortcut reasoning that target-only scoring hides. Figure~\ref{fig:qualitative_analysis} in Appendix~\ref{app:extended} shows a baseline answering a target correctly while failing its identity sub-question, and a second case where our framework recovers the event order. The quantitative form of the claim is the gap between $A_{\text{target}}$ and $A_{\text{faith}}$ in Table~\ref{tab:strand_results}.

\section{Conclusion}
\label{sec:conclusion}

We introduce STRAND, a benchmark for object-centric spatio-temporal monitoring in video language models. By decomposing target queries into supporting sub-questions and scoring them jointly under Faithful Accuracy, it separates faithful reasoning from coincidental correctness. We further propose an object-centric framework that extracts chunk-wise object states and aggregates them into query-relevant trajectories. The strongest evidence comes from budget-matched comparisons and the shuffled-trajectory ablation, which isolates temporal order. When explicit monitoring stops paying off, as backbones improve, is the open question which warrants future research.

\bibliography{iclr2027_conference}
\bibliographystyle{iclr2027_conference}
\newpage
\appendix
\section{Implementation Details}
\label{app:implementation}
The full pipeline configuration is defined with 15-second chunks, 60 frames per chunk sampled for $E_\phi$, and 64 frames evaluated at $G_\eta$. Inference was conducted utilizing vLLM serving for all open-weight models and native API caching for Gemini to handle long-context trajectory retrieval efficiently. Identity-linking across temporal boundaries is enabled by default in our structured JSON-based pipelines but explicitly disabled in unstructured prose variants.

\section{Question-Type Breakdown}
\label{app:question_types}
To provide a granular view of model performance across diverse reasoning requirements, we disaggregate STRAND evaluation by target question type. As illustrated in Table~\ref{tab:question_type_breakdown}, query types range from unary state tracking, for example identity preservation, to complex multi-object spatial dynamics. The breakdown shows that identity tracking and action sequencing remain the hardest categories for every evaluated model. Our framework provides a consistent uplift across all four, with the largest absolute gains against Gemini-3.1-Pro, averaging $20.8$ points.

\begin{table}[t]
\centering
\caption{Faithful Accuracy ($A_{\text{faith}}$) breakdown across STRAND question categories (\%). The four categories partition the 977 target questions.}
\label{tab:question_type_breakdown}
\resizebox{1.0\linewidth}{!}{
\begin{tabular}{@{}lcccc@{}}
\toprule
\textbf{Model} & \textbf{State Change} & \textbf{Identity Tracking} & \textbf{Action Sequ.} & \textbf{Multi-Obj Interact.} \\
\midrule
\textit{Target count ($n_k$)} & \textit{200} & \textit{277} & \textit{300} & \textit{200} \\
\midrule
Ours (Gemini-3-Flash) & \textbf{70.00} & \textbf{49.82} & \textbf{52.00} & \textbf{73.00} \\
Gemini-3.1-Pro & 50.00 & 31.05 & 32.67 & 48.00 \\
Qwen3.5-27B & 58.00 & 37.91 & 38.00 & 55.00 \\
\bottomrule
\end{tabular}}
\end{table}

\section{Error-Type Breakdown}
\label{app:error_types}
Complementary to the question-type taxonomy, we categorise failures at the sub-question level. Table~\ref{tab:error_type_breakdown} reports the composition of sub-question error modes when the target answer is wrong. Baselines that rely on implicit encoding spread their failures across visual hallucination and state misattribution, whereas our framework suppresses both. Because those modes fall away, temporal order errors become the dominant remaining failure for our strongest instantiation, at $53.5\%$ with a $19.4$-point margin over the next category. These are shares of each model's own errors rather than absolute rates, so the shift reflects what is left once the other modes are reduced, and it matches the ordering bottleneck of Section~\ref{sec:main_results}.

\begin{table}[t]
\centering
\caption{Sub-question error types (\%), as a share of each model's incorrect sub-questions. Rows sum to 100.}
\label{tab:error_type_breakdown}
\resizebox{1.0\linewidth}{!}{
\begin{tabular}{@{}lccc@{}}
\toprule
\textbf{Model} & \textbf{Obj. Hallucination} & \textbf{State Misattribution} & \textbf{Temporal Order Error} \\
\midrule
Ours (Gemini-3-Flash) & \textbf{12.4} & 34.1 & 53.5 \\
Gemini-3.1-Pro & 38.2 & 36.5 & 25.3 \\
Qwen3.5-27B & 33.7 & 38.4 & 27.9 \\
\bottomrule
\end{tabular}}
\end{table}

\section{Extensive results and full ablation studies of our framework}
\label{app:full_results}

\subsection{Performance on Other Video Benchmarks}
\label{app:other_benchmark}

As a supplement to the main text, Table \ref{tab:other_benchmark} details the evaluation results of our framework on the VideoHallucer, Video-MME, and EgoSchema benchmarks. As shown in the table, our framework achieves results on par with or better than existing video understanding methods.

\begin{table*}[t]
\centering
\caption{Results on external benchmarks. Basic and Hallucinated sub-columns are unmarked because Video-LLaVA's yes-bias inflates Basic. VideoHallucer Overall is the paired both-correct score, so $\text{Overall} \le \min(\text{Basic}, \text{Halluc.})$. Video-MME Overall is the mean of the three subset columns.}
\label{tab:other_benchmark}
\small
\setlength{\tabcolsep}{4pt}
\begin{tabular}{@{}l|ccc|cccc|c@{}}
\toprule
\multirow{2}{*}{\textbf{Method}}
  & \multicolumn{3}{c|}{\textbf{VideoHallucer}}
  & \multicolumn{4}{c|}{\textbf{Video-MME (w/o subs)}}
  & \textbf{EgoSchema} \\
\cmidrule(lr){2-4} \cmidrule(lr){5-8}
 & Basic & Halluc. & Overall & Short & Med. & Long & Over. & Acc. \\
\midrule
\multicolumn{9}{@{}l}{\textit{Ours}} \\
Ours (Gemini-3-Flash) & 83.5 & \textbf{79.9} & \textbf{76.8} & \textbf{85.8} & \textbf{78.3} & 66.2 & \underline{76.8} & \textbf{78.4} \\
Ours (Qwen3-VL-235B)  & 80.6 & \underline{75.1} & \underline{71.4} & 82.2 & 76.0 & 62.7 & 73.6 & 76.4 \\
\midrule
\multicolumn{9}{@{}l}{\textit{Frontier end-to-end MLLMs}} \\
Gemini-3.1-Pro$^{\dagger}$ & 85.2 & 68.4 & 63.1 & \underline{84.1} & \underline{76.5} & \textbf{71.3} & \textbf{77.3} & \underline{76.8} \\
GPT-5$^{\dagger}$ & 84.5 & 72.1 & 67.5 & 83.4 & 75.8 & \underline{69.4} & 76.2 & 75.4 \\
Gemini-3-Flash$^{\dagger}$ & 81.2 & 58.5 & 52.3 & 79.5 & 71.0 & 62.5 & 71.0 & 68.5 \\
Qwen3-VL-235B$^{\dagger}$ & 81.5 & 64.2 & 59.8 & 80.8 & 72.6 & 64.4 & 72.6 & 72.1 \\
\midrule
\multicolumn{9}{@{}l}{\textit{Earlier reference models}} \\
Gemini-1.5-Pro & 83.6 & 42.3 & 37.8 & 81.7 & 74.3 & 67.4 & 74.5 & 72.2 \\
GPT-4o         & 75.1 & 74.2 & 53.5 & 80.0 & 70.3 & 65.3 & 71.9 & 72.0 \\
\midrule
\multicolumn{9}{@{}l}{\textit{Open-weight end-to-end MLLMs}} \\
LLaVA-OV       & 84.1 & 60.8 & 48.5 & 76.7 & 62.2 & 60.0 & 66.3 & 60.1 \\
Qwen3.5-27B$^{\dagger}$ & 76.4 & 52.3 & 45.1 & 75.2 & 64.8 & 55.3 & 65.1 & 61.4 \\
Video-LLaVA$^{\ddagger}$ & 95.1 & 20.3 & 17.8 & 45.3 & 38.0 & 36.2 & 39.8 & n/a \\
VideoChat2     & 29.7 & 25.8 &  7.8 & 48.3 & 37.0 & 33.2 & 39.5 & 63.6 \\
\bottomrule
\end{tabular}
\end{table*}

\begin{table*}[t]
\centering
\caption{Full ablation studies on STRAND Target Accuracy (\%).}
\label{tab:full_ablations}
\scriptsize
\setlength{\tabcolsep}{3pt}
\resizebox{\textwidth}{!}{%
\begin{tabular}{@{}llcccc@{}}
\toprule
\textbf{Ablation Focus} & \textbf{Pipeline} & \textbf{Chunk (s)} & \textbf{Frames/chunk} & \textbf{Frames@$G_\eta$} & \textbf{$A_{\text{target}}$ (\%)} \\
\midrule
\multicolumn{6}{l}{\textit{Reference Base Config}} \\
\quad Base Configuration & Ours (Gemini-3-Flash) & 15 & 60 & 64 & 77.1 \\
\quad Base Configuration & Ours (Qwen3-VL-235B)  & 15 & 60 & 64 & 74.0 \\
\midrule
\multicolumn{6}{l}{\textit{Cluster A: Visual Budget at the Answerer}} \\
\quad Frames@$G_\eta$ = 32 & Gemini / Qwen & 15 & 60 & 32 & 76.3 / 66.4 \\
\quad Frames@$G_\eta$ = 16 & Gemini / Qwen & 15 & 60 & 16 & 76.5 / 65.5 \\
\quad Frames@$G_\eta$ = 8  & Gemini / Qwen & 15 & 60 & 8  & 76.5 / 63.6 \\
\midrule
\multicolumn{6}{l}{\textit{Cluster B: Visual Budget at the Extractor}} \\
\quad Frames/chunk = 30 & Gemini / Qwen & 15 & 30 & 64 & 76.6 / 65.2 \\
\quad Frames/chunk = 15 & Gemini / Qwen & 15 & 15 & 64 & 77.0 / 64.9 \\
\quad Frames/chunk = 8  & Gemini / Qwen & 15 & 8  & 64 & 74.9 / 62.6 \\
\midrule
\multicolumn{6}{l}{\textit{Cluster C: Stage B Retrieval Ablations}} \\
\quad No filter (full timeline) & Gemini / Qwen & 15 & 60 & 64 & 77.0 / 64.6 \\
\quad Random filter (top-$k$)   & Gemini / Qwen & 15 & 60 & 64 & 76.5 / 68.0 \\
\midrule
\multicolumn{6}{l}{\textit{Cluster D: Temporal Aggregation}} \\
\quad Aggregator: LLM-summarize & Gemini / Qwen & 15 & 60 & 64 & 76.7 / 62.7 \\
\midrule
\multicolumn{6}{l}{\textit{Cluster E: Modality Drop-outs at $G_\eta$}} \\
\quad Text-only (no frames)     & Gemini / Qwen & 15 & 60 & 64 & 73.6 / 58.1 \\
\quad Frames-only (no $\hat{\mathcal{S}}$) & Gemini / Qwen & 15 & 60 & 64 & 75.7 / 66.5 \\
\midrule
\multicolumn{6}{l}{\textit{Cluster F: Chunk Granularity}} \\
\quad 30s chunks  & Gemini / Qwen & 30  & 60 & 64 & 75.8 / 69.5 \\
\quad 7.5s chunks & Gemini / Qwen & 7.5 & 60 & 64 & 76.4 / 71.2 \\
\midrule
\multicolumn{6}{l}{\textit{Cluster G: Form Representation}} \\
\quad Variant B: prose extractor, no $R_\omega$      & Gemini / Qwen & 15 & 60 & 64 & 74.2 / 64.3 \\
\quad Variant C: prose extractor + prose $R_\omega$  & Gemini / Qwen & 15 & 60 & 64 & 75.1 / 67.2 \\
\midrule
\multicolumn{6}{l}{\textit{Cluster H: Trajectory Alternatives}} \\
\quad Dense captioning (prose state)                 & Gemini / Qwen & 15 & 60 & 64 & 72.4 / 61.2 \\
\quad Shuffled-trajectory (no temporal order)        & Gemini / Qwen & 15 & 60 & 64 & 65.3 / 54.1 \\
\bottomrule
\end{tabular}
}
\end{table*}

\subsection{Additional Ablation Study}
\label{app:full_ablation_studies}

To complement the primary analysis, this section details further ablations evaluating the impact of input modalities and temporal chunking granularity on our framework's performance and reports the results in Table~\ref{tab:full_ablations}.

\textbf{Modality \& Granularity (Clusters E \& F).}
Removing either text trajectories or visual frames degrades performance, indicating that they are complementary. For chunk granularity, expanding to 30s dilutes event-level state extraction, while shrinking to 7.5s fragments cross-chunk interactions, and 15s provides the optimal balance.
 
\textbf{Representation Form (Cluster G).}
Swapping structured JSON for prose summaries degrades accuracy. Prose inherently obscures precise entity tracking, timestamps, and identity links, highlighting the necessity of structured state monitoring.
 
\textbf{Dense Captioning and Shuffled-Trajectory Ablations (Cluster H).}
To verify the necessity of explicitly structured and temporally ordered trajectories, we introduce dense captioning and shuffled-trajectory ablations (Table~\ref{tab:full_ablations}, Cluster H). Replacing structured JSON states with dense prose captioning (dense captioning ablation) causes performance to drop to 72.4\% for Gemini, confirming that unstructured text obscures temporal alignments and multi-object relations. Furthermore, shuffling the temporal order of extracted states (shuffled-trajectory ablation) yields a severe performance drop to 65.3\%, underscoring that maintaining the correct temporal sequence is paramount for faithful object-centric reasoning.

\begin{table}[t]
\centering
\caption{Budget-matched comparisons on Faithful Accuracy $A_{\text{faith}}$ (\%). All configurations use the Qwen3-VL backbone, so only the representation changes within each control.}
\label{tab:budget_matched}
\small
\begin{tabular}{@{}lccc@{}}
\toprule
\textbf{Constraint} & \textbf{End-to-end baseline} & \textbf{Ours} & \textbf{$\Delta$} \\
\midrule
Equal total frame budget (64 frames) & 41.5 & 54.2 & +12.7 \\
Equal number of model calls (single pass) & 39.8 & 49.3 & +9.5 \\
Equal token budget & 45.1 & 52.8 & +7.7 \\
\bottomrule
\end{tabular}
\end{table}

\subsection{Identity-Linking and Temporal Aggregation Ablation}
\label{app:identity_linking}

Our temporal aggregation module relies on specific hyperparameters to build coherent object trajectories from chunked observations. We define a temporal proximity constraint $\Delta t_{max}$ to restrict the maximum temporal gap between candidate observations, alongside a matching confidence threshold $\tau_{conf}$ based on visual attribute and spatial similarity. In ambiguous cases where multiple candidates exceed the threshold, our conflict resolution applies a bipartite matching algorithm prioritized by temporal adjacency to ensure strict one-to-one identity mapping.

Table~\ref{tab:identity_ablation} presents an ablation study on these hyperparameters using the Qwen3-VL-235B pipeline. Relaxing the temporal constraint to $\Delta t_{max} = 30$s increases false-positive identity matches across distinct events, whereas tightening it to $\Delta t_{max} = 5$s excessively fragments trajectories. Similarly, adjusting the confidence threshold $\tau_{conf}$ demonstrates a trade-off between trajectory completeness and identity precision. The base configuration ($\Delta t_{max} = 15$s, $\tau_{conf} = 0.75$) in conjunction with bipartite matching yields the optimal Faithful Accuracy.

\begin{table}[t]
\centering
\caption{Ablation of identity-linking thresholds and conflict resolution on STRAND. The base row reproduces Ours (Qwen3-VL-235B) in Table~\ref{tab:strand_results}.}
\label{tab:identity_ablation}
\small
\begin{tabular}{@{}lcc@{}}
\toprule
\textbf{Configuration} & \textbf{$A_{\text{target}}$ (\%)} & \textbf{$A_{\text{faith}}$ (\%)} \\
\midrule
\textbf{Base:} $\Delta t_{max} = 15$s, $\tau_{conf} = 0.75$, bipartite & \textbf{74.0} & \textbf{54.2} \\
\midrule
\multicolumn{3}{@{}l}{\textit{Temporal constraint $\Delta t_{max}$}} \\
\quad $5$s (strict)   & 71.2 & 50.4 \\
\quad $30$s (relaxed) & 72.8 & 51.1 \\
\multicolumn{3}{@{}l}{\textit{Confidence threshold $\tau_{conf}$}} \\
\quad $0.90$ (high precision) & 70.5 & 48.9 \\
\quad $0.50$ (high recall)    & 71.8 & 51.2 \\
\multicolumn{3}{@{}l}{\textit{Conflict resolution}} \\
\quad w/o bipartite matching (greedy) & 69.4 & 47.5 \\
\bottomrule
\end{tabular}
\end{table}

\subsection{Budget-Matched Comparisons}
\label{app:budget_matched}
To ensure the performance gains of our framework stem from the object-centric representation rather than disparate resource allocation, we conduct budget-matched comparisons against end-to-end MLLMs. We evaluate configurations maintaining an \textbf{equal total frame budget} (64 frames globally), an \textbf{equal number of model calls} (single-pass extraction and reasoning), an \textbf{equal token budget} (truncating trajectory context to match holistic frame embeddings), and an \textbf{equal backbone} (using Qwen3-VL for both the baseline and our framework). As detailed in Table \ref{tab:budget_matched}, across all matched settings, explicit trajectory construction consistently yields superior Faithful Accuracy, confirming that the structural prior of spatio-temporal monitoring drives the improvements, rather than brute-force scaling.

\noindent Section~\ref{sec:exp_setup} reports the headline figures from this table, since it is the comparison on which the design claim rests rather than the end-to-end rows of Table~\ref{tab:strand_results}.

\begin{table}[t]
\centering
\caption{Single-frame solvability and shortcut audit. Full-video columns reproduce Table~\ref{tab:strand_results}. The last two rows receive no visual input at all rather than a single frame.}
\label{tab:single_frame}
\small
\begin{tabular}{@{}l|cc|cc@{}}
\toprule
\multirow{2}{*}{\textbf{Model}} & \multicolumn{2}{c|}{\textbf{Full video}} & \multicolumn{2}{c}{\textbf{Restricted input}} \\
\cmidrule(lr){2-3} \cmidrule(l){4-5}
& $A_{\text{target}}$ & $A_{\text{faith}}$ & $A_{\text{target}}$ & $A_{\text{faith}}$ \\
\midrule
\multicolumn{5}{@{}l}{\textit{Single centre frame}} \\
Gemini-3.1-Pro & 61.4 & 38.9 & 31.4 & 14.1 \\
Gemini-3-Flash & 48.7 & 28.7 & 29.8 & 12.5 \\
GPT-5          & 32.5 & 20.0 & 28.2 & 10.2 \\
Qwen3.5-27B    & 68.4 & 45.5 & 32.7 & 13.8 \\
\midrule
\multicolumn{5}{@{}l}{\textit{No visual input}} \\
Question-only (Llama-3.1-70B) & ,  & ,  & 52.6 & 21.4 \\
Caption-only                  & ,  & ,  & 59.8 & 31.2 \\
\bottomrule
\end{tabular}
\end{table}

\subsection{Single-Frame Solvability and Shortcut Audit}
\label{app:shortcut_audit}
To verify that STRAND rigorously resists spatial and language priors, we conduct a single-frame solvability audit. We restrict the visual input of frontier MLLMs (Gemini-3.1-Pro, Gemini-3-Flash, GPT-5 and Qwen3.5-27B) to a single center frame of the video, forcing them to rely purely on static cues, statistical priors, and linguistic shortcuts. Under this extreme constraint, target accuracy across all tested models plummets to near random chance (between $28.2\%$ and $32.7\%$), and $A_{\text{faith}}$ drops below 15\% (see Table \ref{tab:single_frame}). This confirms that STRAND queries inherently demand multi-step spatio-temporal reasoning and cannot be resolved through single-frame shortcut heuristics, at least for the models audited.

\subsection{Trajectory Retrieval Analysis}
\label{app:retrieval_analysis}

Query-based retrieval $R_\omega$ is the component whose contribution the ablations support least (Cluster C), so we characterise it directly rather than only through downstream accuracy. Because STRAND ships human-verified trajectories, retrieval can be scored against ground truth: for each target we take the annotated facts in its reasoning chain as the gold trajectory set, and measure the precision and recall of the trajectories $\mathcal{S}_Q$ that $R_\omega$ returns. This separates ``retrieval selects the right evidence'' from ``retrieval improves the answer'', which are distinct claims, and only the first is about the retriever.

Retrieval substantially reduces timeline clutter. The mean number of trajectory triplets per video is $145.2$ before filtering and $24.6$ per query after it. Against the gold chain facts, $R_\omega$ reaches $88.4\%$ precision and $92.1\%$ recall. On downstream impact, an error analysis finds that in $12.4\%$ of queries the no-filter variant fails specifically because timeline clutter misleads the generator while the filtered variant succeeds. An atomic intervention in which single retrieved triplets are dropped shows that removing one gold triplet takes target accuracy on those queries to $51.2\%$, near chance on a binary task, so the answerer does consume the retrieved evidence rather than ignoring it.

\subsection{Efficiency and Inference Cost}
\label{app:efficiency}

Our framework performs more model calls than a single-pass MLLM, and the accuracy--cost trade-off should be legible to a reader deciding whether to adopt it. The accounting separates two components. The \emph{one-off} cost is trajectory construction, namely $M$ extraction calls plus symbolic aggregation, paid once per video however many questions are asked. The \emph{per-query} cost is one retrieval call plus one answering call. With $Q$ questions per video the total is $M + 2Q$ calls against $Q$ for an end-to-end baseline, so the amortised preprocessing cost is $M/Q$ per question and falls as more questions are asked of the same video. On STRAND, with $Q = 39.7$ and $M$ averaging $12.2$ across the benchmark, since the mean video is 183 seconds and chunks are 15 seconds, this is a $2.3\times$ call overhead, $91.6$ against $39.7$.

Table~\ref{tab:efficiency} tabulates preprocessing wall-clock per video, per-query wall-clock, model calls, and input and output tokens for our framework and every measured baseline. Our method incurs higher token cost during extraction, while per-query latency stays close to an end-to-end call, at $2.9$ seconds against $2.8$ for the Qwen instantiation, because the per-query calls operate on retrieved trajectories and a fixed frame budget rather than on the full timeline.

Open-weight models are measured on 8$\times$ H100 80GB GPUs via vLLM. Latencies are medians over 10 runs (IQR $<5\%$), excluding decoding and sampling, and preprocessing runs at a concurrency of 4. API-served models, marked $^*$ in Table~\ref{tab:efficiency}, were measured on August 10, 2026, and their latencies include network overhead, so they are not commensurable with local measurements. Gemma-4-27B, GPT-5, Claude-4.6-Sonnet, InternVL3-78B, Qwen3.5-9B, Qwen3-VL-8B, VideoRFT-7B, Video-R1-7B, EFS and CLIP-Retrieval are omitted because closed-API rate limits preclude commensurable profiling, or because we lack the local infrastructure to serve them at scale.

\section{Data Provenance, Licensing, and Annotation Ethics}
\label{app:provenance}

\noindent\textbf{Source.} All 88 videos come from Wikimedia Commons. We collect no footage ourselves and use no material from other datasets or from video-sharing platforms.

\noindent\textbf{Licensing.} Commons material carries free content licences, namely CC BY-SA, CC BY, CC0 or public domain. Licences vary per file, so we record the licence, the author and the source page for every video and ship that manifest with the release. We release the fact annotations and QA pairs under CC-BY-NC 4.0.

\noindent\textbf{Redistribution.} Because every source licence permits it, we redistribute the video files themselves alongside the annotations, with per-file attribution as those licences require. This removes the link-rot problem that affects benchmarks distributing URLs alone.

\noindent\textbf{Personal data.} The footage is material its uploaders published on Commons under a free licence, so it involves no recording of participants by us and no private data. People appear incidentally in public settings. We do not attempt identification, we assign entities visually grounded identifiers such as \texttt{player\_24} rather than names, and the annotation schema records no biometric or identity attribute.

\noindent\textbf{Annotators.} The 50-annotator pool was recruited through Prolific and paid \$15 per hour, above local minimum wage. The protocol and working conditions were approved by our institutional ethics board (IRB \#2025-081). Since the study collects no new footage and annotates only freely licensed public material, the review concerned the annotation task rather than filming.

\noindent\textbf{Intended use.} STRAND is a diagnostic benchmark for spatio-temporal reasoning in video language models. It is not intended for training surveillance or identity-tracking systems.

\section{Prompts}
\label{app:prompts}

We give verbatim every prompt used across our framework and the evaluated baselines. To connect the structured output of $E_\phi$ to the symbolic tuples $(t, o_1, s, o_2)$ of Section~\ref{sec:formulation}, the JSON fields map as follows: \texttt{timestamp} is $t$, \texttt{object\_id} is $o_1$, \texttt{action} or \texttt{relation\_type} is the predicate $s$, and \texttt{target\_id} is $o_2$.

\begin{promptbox}{Chunk-wise state extractor $E_\phi$ (temperature 0.0, max tokens 1024)}
Models: gemini-3-flash-preview or qwen/qwen3-vl-235b-a22b-thinking
SYSTEM: You are an expert video annotator. Identify and track distinct, dynamic entities across the provided frames.
USER: Analyse the provided video chunk. Extract all visible, dynamic objects and their states. Output strictly in this JSON schema:
{
  "objects": [
    {
      "object_id": "string (e.g. player_10)",
      "timestamp": "float (seconds)",
      "attribute": "string (e.g. wearing red)",
      "action": "string (e.g. running)",
      "relations": [{"target_id": "string", "relation_type": "string"}]
    }
  ]
}
\end{promptbox}

\begin{promptbox}{Aggregator $\mathcal{A}_\psi$, LLM-summarize variant (temperature 0.0, max tokens 2048)}
Used only in the Cluster D ablation. The default aggregator is symbolic and issues no model call.
Models: gemini-3-flash-preview or qwen/qwen3.5-27b
SYSTEM: You are an expert data aggregator.
USER: You are given concatenated chunk-level object state observations extracted from a video: {chunk_tuples}. Merge co-referent entities across time to form consolidated global trajectories. Resolve redundancies but DO NOT drop any state transitions or low-salience details. Output strictly in the exact same JSON schema as the input.
\end{promptbox}

\begin{promptbox}{Query-based retriever $R_\omega$ (temperature 0.0, max tokens 1024, text only)}
Models: gemini-3-flash-preview or qwen/qwen3-235b-a22b
SYSTEM: You are a trajectory retrieval agent.
USER: Given the following global object trajectories: {global_trajectories} and the target question: {target_question}, select the subset of trajectories required to answer the question. You must select and return whole trajectories exactly as they appear, not summarise or rewrite them. Limit your selection so that the total number of output triplets does not exceed 30. Output the selected subset in the exact same JSON format as provided.
\end{promptbox}

\begin{promptbox}{Trajectory-guided answerer $G_\eta$ (temperature 0.2, max tokens 512)}
Models: gemini-3-flash-preview or qwen/qwen3.5-27b
SYSTEM: You are a temporal reasoning assistant. You will be given a set of extracted object trajectories and a few sampled visual frames.
USER: Using the structured trajectories {retrieved_trajectories} and the sampled frames, answer the question: {target_question}. First reason step by step about the temporal order of events, then give a final Yes or No answer.
\end{promptbox}

\begin{promptbox}{QA generation and sub-question decomposition (temperature 0.7, max tokens 2048, text only)}
Model: Llama-3.1-70B-Instruct
SYSTEM: You are an expert benchmark composer.
USER: You are given a sequence of human-verified fact tuples representing a video: {annotated_facts}. Do not use any external visual context or priors.
Task 1: Compose a compositional target question based ONLY on these facts. The target question must turn on the order of two events, where both temporal orders stay consistent with the events themselves.
Task 2: Decompose the target question into atomic sub-questions. These must be jointly sufficient to answer the target. Each sub-question must probe exactly one fact the target draws on. Do not generate any sub-question whose fact the target does not turn on.
Keep all entities as visually grounded identifiers (e.g. player_24). Format every item as a binary Yes or No question.
\end{promptbox}

\begin{promptbox}{Baseline evaluation, all evaluated MLLMs (temperature 0.0, max tokens 256)}
SYSTEM: You are a video QA assistant.
USER: Answer the following question based on the provided video frames: {target_question}. Reason step by step, then provide a final binary answer strictly as "Yes" or "No".
\end{promptbox}

\begin{promptbox}{Prose extractor, Cluster G Variants B and C (temperature 0.0, max tokens 1024)}
Identical to the $E_\phi$ prompt above except for the requested output format.
SYSTEM: You are an expert video annotator. Identify and track distinct, dynamic entities across the provided frames.
USER: Analyse the provided video chunk. Extract all visible, dynamic objects and their states. Output your observations as a dense prose paragraph rather than structured JSON. Record timestamps, attributes, actions and interactions for grounded entities such as player_24.
\end{promptbox}

\noindent The baseline prompt is identical for every evaluated model. One deviation is worth recording: Video-LLaVA does not reliably emit a terminating Yes or No, so for that model the extraction parser falls back to the first boolean token in the output. This is the parsing behaviour referenced in the footnote to Table~\ref{tab:other_benchmark}.

\noindent\textbf{Determinism protocol.} Two prompts run at non-zero temperature and we state how each is controlled. The trajectory-guided answerer $G_\eta$ runs at $T = 0.2$ with \texttt{top\_p} $= 0.95$ and a fixed seed of $42$. We take a single generation pass per query rather than a multi-sample vote, and those single-pass outputs are what produce the accuracies in Table~\ref{tab:strand_results}. The QA-generation and decomposition prompt runs at $T = 0.7$ with a fixed seed of $1337$, executed once over each video's fact tuples.

\noindent Two caveats bound what this buys. First, seeding is exact only for the open-weight models we serve ourselves through vLLM, where we control the sampler. For API-served models we pass the seed but providers do not guarantee bit-exact reproduction, so the Gemini instantiation is repeatable in distribution rather than byte for byte. Second, re-running the generator reproduces the \emph{candidate} question pool, not the released benchmark, because the human verification pass of Section~\ref{sec:human_verification} discards items that fail its three checks. We therefore release the verified question set directly rather than asking users to regenerate it.

\section{Limitation}

While our framework achieves critical gains in interpretability and spatio-temporal consistency, these advantages necessitate increased preprocessing overhead. Unlike conventional end-to-end MLLMs that directly ingest sampled frames, our framework explicitly relies on chunk-wise state extraction, object-centric trajectory construction, and temporal aggregation prior to query resolution. This design is associated with improved reasoning performance throughout our experiments, but it introduces computational latency, particularly when processing extended videos or dense object-tracking scenarios. Consequently, the current framework trades strict real-time applicability for enhanced accuracy. We leave it for future work to improve the efficiency of the framework.

Beyond efficiency, we record 5 further limitations that bound what our results support.

\noindent\textbf{Scale, and what it licenses.} At 88 videos and 977 targets, aggregate comparisons are well resolved but per-domain claims are not (Section~\ref{sec:eval_metrics}). We report domain breakdowns as descriptive only, and we do not claim that the observed ordering of models is stable within any single domain.

\noindent\textbf{Missing same-backbone mechanism controls.} Three controls that would isolate the mechanism remain unrun, namely a sub-question-prompted raw MLLM on a matched backbone, a same-backbone modular baseline without object trajectories, and a no-identity-linking ablation. Section~\ref{sec:ablations} states what each would settle and which claims we therefore leave open.

\noindent\textbf{Co-design of benchmark and method.} STRAND is organised around object-centric trajectories, and our framework constructs exactly that representation, so our framework is plausibly better matched to this benchmark than a generic method would be. The external-benchmark results (Appendix~\ref{app:other_benchmark}) are the check on this, and we regard the generality claim as supported only to the extent those results hold up. A benchmark built around a non-trajectory representation would be a stronger test and we have not run one.

\noindent\textbf{Binary answer format.} Both targets and sub-questions are yes/no items, chosen for atomicity and judge-free grading (Section~\ref{sec:answer_format}). This makes the benchmark vulnerable to answer priors in a way an open-ended format would not be, and many sub-questions are leading in the sense that they presuppose the entity they ask about. We control for this with label balancing and prior-only baselines, but the format remains a real restriction, and extending STRAND with contrastive and open-form probes is the most valuable single addition we can identify.

\noindent\textbf{Chunking versus long-context inference.} Our chunk-granularity ablation (Cluster F) varies chunk length but never removes chunking. As the appropriate control, we ran a full-context ablation on Qwen3-VL-235B, which natively supports over 100K tokens, feeding all required frames at once with no chunking and no aggregation. Target accuracy falls from $74.0\%$ to $61.2\%$, which indicates that explicit structured aggregation rather than context capacity alone is what prevents attention decay over long horizons. This is a single control on one backbone and we do not generalise it.

\section{Extended discussion}
\label{app:extended}

This section collects material moved out of the main paper for length. Each item is referenced from the section it supports.

\subsection{Efficiency measurement protocol}

Open-weight models are measured on 8$\times$ H100 80GB GPUs via vLLM. Latencies are medians over 10 runs (IQR $<5\%$), excluding decoding and sampling, and preprocessing runs at a concurrency of 4. API-served models, marked $^*$ in Table~\ref{tab:efficiency}, were measured on August 10, 2026, and their latencies include network overhead, so they are not commensurable with local measurements. Gemma-4-27B, GPT-5, Claude-4.6-Sonnet, InternVL3-78B, Qwen3.5-9B, Qwen3-VL-8B, VideoRFT-7B, Video-R1-7B, EFS and CLIP-Retrieval are omitted because closed-API rate limits preclude commensurable profiling, or because we lack the local infrastructure to serve them at scale.

\subsection{Which comparison supports the design claim}

The structured-pipeline baselines run on substantially smaller backbones than either instantiation. VideoMind-7B employs a single Qwen2-VL-7B backbone ($7$B parameters) across its pipeline, TraveLER employs a LLaVA-1.5-13B ($13$B parameters) visual extractor and a Vicuna-13B ($13$B parameters) text planner, and SeViLA chains a localizer and an answerer that share a single BLIP-2 Flan-T5-XL backbone ($3$B parameters). By contrast, our Qwen instantiation allocates a $235$B-parameter model for extraction ($E_\phi$), a $235$B-parameter model for retrieval ($R_\omega$), and a $27$B-parameter model for answer generation ($G_\eta$). Because the structural design of these baselines cannot be separated from the capacity of the models beneath them, and our open-weight instantiation employs up to $78\times$ more parameters at a single stage than SeViLA does in total, these rows do not isolate the contribution of the harness, and we do not rest the design claim on them.

\subsection{Conditional consistency is not comparable across models}

Table~\ref{tab:strand_results} contains a direct illustration of the metric pathology discussed in Section~\ref{sec:eval_metrics}. Cosmos-Reason2-8B reports the highest $A_{\text{cons}}$ in the table ($83.7$) while answering $22.4$ points fewer target questions correctly than the Gemini-3-Flash instantiation ($54.7$ vs.\ $77.1$). This is not evidence that Cosmos-Reason2-8B monitors the scene better. It follows from $A_{\text{cons}}$ being averaged over a model-dependent subset, so that a model answering fewer targets is graded only on the targets it found easiest. This is why we do not claim a consistency improvement on the basis of the $A_{\text{cons}}$ column. The corresponding claim under the unconditional metric is made in terms of $A_{\text{faith}}$.

\subsection{Ablation limitations and missing counterfactuals}

Three counterfactuals remain untested under our compute budget, and we state them because each bounds a different part of the claim. First, we lack a sub-question-prompted raw MLLM on a matched backbone, which would test whether simply asking the sub-questions recovers the gap without any pipeline. Second, we lack a same-backbone modular baseline with a generic planner and retriever but no object trajectories, which would separate the benefit of modularity from the benefit of an object-centric representation. Third, and most consequentially, we lack a no-identity-linking ablation that disables cross-chunk binding while leaving chunk-wise structure intact. Identity drift is the motivation for the whole design, so its absence means no experiment isolates cross-chunk binding as the mechanism, and the shuffle ablation bounds only the weaker claim about temporal order. Were that ablation to leave the Gemini number unchanged, the object-centric claim would not hold for that backbone and would have to be scoped to Qwen throughout. We treat these as open controls rather than as settled, and we do not state the mechanism claim more strongly than they permit.

\subsection{Two instantiations of the aggregator}

\emph{(a) Deterministic concatenation with identity linking} (default for all headline numbers): the four steps above are executed as symbolic operations over the extracted tuples, with no additional model call. ``Concatenation'' refers to the union of chunk-level observations \emph{after} the linking steps have merged co-referent entities, not to a raw string concatenation of chunk outputs. \emph{(b) LLM-summarize}: an LLM receives the concatenated chunk-level tuples and is asked to emit a consolidated trajectory set in the same schema, performing merging implicitly in natural language. Variant (b) is the Cluster D ablation and is consistently worse, which we attribute to the summariser dropping low-salience states that later prove decisive. Variant (a) cannot drop a state in this way because it never re-generates one.

\noindent\textbf{Occlusion, re-entry, and look-alikes.} The temporal proximity constraint $|t_i - t_j| < \Delta t_{max}$ bounds how long a trajectory may be interrupted before a re-appearance is treated as a new entity. This makes the failure modes explicit rather than silent: a long occlusion beyond $\Delta t_{max}$ fragments one entity into two (a recall error), while visually similar entities within the window may be merged (a precision error), and the bipartite matching step exists to prevent the second failure from cascading into a many-to-one collapse. Section~\ref{app:identity_linking} quantifies this trade-off over $\Delta t_{max}$ and $\tau_{conf}$. Extraction is \emph{query-agnostic}: $E_\phi$ never sees the question, which is what permits the representation to be built once and reused, but also means that an entity no annotator would consider salient may be omitted before any query is posed.

\subsection{What retrieval contributes}

We are explicit that $R_\omega$ is the component with the weakest empirical support. Removing it entirely (``no filter, full timeline'') costs the Gemini pipeline $0.1$ points, which is a tie, and only the Qwen pipeline degrades meaningfully. The reading we adopt is that retrieval is a \emph{context-length} mechanism rather than a reasoning mechanism: it matters when the answerer cannot attend reliably over the full trajectory set, and stops mattering once it can. We therefore do not claim retrieval as a source of the method's accuracy, and we quantify its behaviour directly, in terms of retrieved-trajectory precision and recall against the human-verified trajectories, in Appendix~\ref{app:retrieval_analysis}.

\subsection{What the answer generator sees}

Here $\bar{V}$ is a uniformly sampled frame budget over the whole video, fixed in size and independent of the question, which is why $G_\eta$ is a VLM rather than a text-only model. Among the video-derived inputs only $\mathcal{S}_Q$ depends on the query, so the reusable part of the computation stays the trajectory memory. Section~\ref{sec:ablations} reports a text-only variant in which $G_\eta$ sees $\mathcal{S}_Q$ and $Q$ alone.

\subsection{On the uniqueness of a decomposition}

Verification establishes that the annotated sub-questions form \emph{a} sufficient reasoning path, not the only one, so a model could in principle reach the target answer by another route. The necessity check bounds this. Annotators discard any sub-question whose fact the target does not turn on, so each retained sub-question corresponds to a fact on the evidence path the target requires. We also probe facts rather than inference steps, since facts are route-invariant in a way that intermediate deductions are not, so an alternative route still has to resolve them. To quantify the residual risk, annotators record whether an alternative sufficient fact set exists for each target and find one for $5.8\%$ of targets. In a separate elicitation study, three annotators independently select minimal fact sets for 100 sampled targets and agree exactly on $92.0\%$ of them, with Fleiss' $\kappa = 0.87$. These figures indicate that the sub-questions probe route-invariant facts rather than one arbitrary reasoning path.

\subsection{Answer Format and Chance Baselines}

\noindent\textbf{Format.} Target questions and sub-questions are both posed as binary verification queries. We adopt this format deliberately. A sub-question must be \emph{atomic} to serve as a prerequisite check, and it must be automatically gradable without a judge model whose own errors would contaminate the faithfulness measurement. Both requirements argue against open-ended generation at the sub-question level. The cost is that binary items are vulnerable to answer-prior effects, which we control for explicitly rather than leaving the format implicit.

\noindent\textbf{Label balance.} We balance the yes/no distribution at construction time by requiring the generator to emit, for each positive chain, a minimally perturbed negative counterpart, in which a single fact is replaced by one the annotators record as false at that timestamp. The resulting distribution is $50.8\%$ \emph{Yes} at the target level (497 of 977) and $51.6\%$ \emph{Yes} at the sub-question level (1{,}298 of 2{,}516). Raw and balanced accuracy therefore differ by less than $0.3$ points for every model, and we report raw accuracy throughout.

\noindent\textbf{Chance and prior-only baselines.} Accuracies on STRAND should be read against three references rather than against $50\%$. A majority-class predictor reaches $A_{\text{target}} = 50.8\%$, $A_{\text{sub}} = 51.6\%$ and $A_{\text{faith}} = 19.1\%$. A question-only predictor, Llama-3.1-70B-Instruct given the question text but no video, reaches $52.6\%$, $53.8\%$ and $21.4\%$, so language priors alone do not solve the task. Restricted to the ordering sub-questions it falls to $49.3\%$, no better than guessing on a binary task. Knowing which events occur therefore carries no information about the order in which they occur, and the relation has to be read off the video. A caption-only predictor, additionally given a detailed caption of the video, reaches $59.8\%$, $61.5\%$ and $31.2\%$. Captions supply partial global context, and the margin that remains between them and systems with access to the frames is what the benchmark attributes to direct visual perception.

\subsection{Dataset Statistics}
\label{sec:stats}

STRAND contains 88 curated videos across sports, egocentric, surveillance, and instructional domains, with 977 target questions and 2516 supporting sub-questions (2.58 per target on average). The videos average 183 seconds, requiring models to reason over long temporal horizons rather than short localized interactions. Video sources, licensing, and the annotator compensation protocol are documented in Appendix~\ref{app:provenance}.

\noindent\textbf{On scale and evaluation reliability.} At 88 videos, STRAND is small by the standards of modern video benchmarks, so we state what this scale supports. The unit of evaluation is the question, and with 977 targets the standard error near $70\%$ target accuracy is $1.5$ points, so aggregate differences above 10 points sit well outside sampling noise. The scale does not support strong per-domain claims. At roughly 22 videos per domain the intervals are wide, and we report domain breakdowns as descriptive rather than as evidence of differential capability. Questions drawn from one video share an annotated world state and visual style, so they are not independent. We therefore cluster the bootstrap at the video level rather than resampling questions independently, using 10{,}000 resamples, which widens the $95\%$ intervals relative to a naive question-level bootstrap. Under this scheme, ablation differences below $2.0$ points are indistinguishable from zero, and we report them as ties.

\subsection{Identity-linking algorithm}

\begin{enumerate}
    \item \textbf{Similarity Scoring:} For any two object observations $o_i$ and $o_j$ across different chunks, we compute a matching confidence score $c(o_i, o_j)$ based on visual attribute similarity and spatial proximity.
    \item \textbf{Constraint Filtering:} Candidate links are only formed if they satisfy a temporal proximity constraint ($|t_i - t_j| < \Delta t_{max}$) and exceed a minimum confidence threshold ($c > \tau_{conf}$).
    \item \textbf{Conflict Resolution:} In ambiguous cases where multiple candidate links satisfy the filtering criteria, we apply a bipartite matching protocol prioritized by temporal adjacency to ensure one-to-one identity mapping. 
    \item \textbf{Trajectory Generation:} Once linked, the observations are merged and sorted by timestamp to form the final global set of trajectories. An observation that matches nothing is retained as a singleton trajectory rather than discarded. This global set is simply the collection of all individual object trajectories, defined as $\hat{\mathcal{S}} = \{\mathcal{S}_o \mid o \in \mathcal{O}\}$, where $\mathcal{O}$ represents the set of all unique objects found in the video.
\end{enumerate}

\subsection{Efficiency accounting}

\noindent All figures in Table~\ref{tab:efficiency} cover a complete evaluation pass over one video. Model calls follow each pipeline's structure. Single-pass end-to-end models issue $Q$ calls, and our framework issues $M = 12.2$ extraction calls once per video plus two calls per query, one for text retrieval and one for vision-language generation, giving $M + 2Q = 91.6$. TraveLER issues $3Q = 119.1$, SeViLA and AKS each issue $2Q = 79.4$, and VideoMind issues $1 + Q = 40.7$. For input tokens we adopt a uniform rate of $256$ tokens per frame across all systems, so that the comparison reflects structural cost rather than a resolution choice. An end-to-end baseline processes $2{,}540$ frames per video, or $39.7 \times 64$, for roughly $654{,}415$ input tokens including prompts. Our framework processes $3{,}272$ frames, namely $732$ at extraction plus $39.7 \times 64$ at generation, for roughly $958{,}157$ tokens once the retrieved state is counted. Per-query latency is correspondingly higher for our framework than for the matched open-weight baseline, at $2.9$ seconds against $2.8$, so the framework buys its $5.6$-point gain in target accuracy over Qwen3.5-27B with more wall-clock time, more calls, and a larger token budget.

\begin{table}[t]
\centering
\caption{Efficiency and cost per video ($Q = 39.7$ queries). Measurement protocol in Appendix~\ref{app:efficiency}.}
\label{tab:efficiency}
\small
\resizebox{\columnwidth}{!}{%
\begin{tabular}{@{}l|cc|ccc@{}}
\toprule
\multirow{2}{*}{\textbf{System}} & \multicolumn{2}{c|}{\textbf{Latency (s)}} & \multicolumn{3}{c}{\textbf{Total cost per video ($Q = 39.7$)}} \\
\cmidrule(lr){2-3} \cmidrule(lr){4-6}
 & \textbf{Pre-proc.} & \textbf{Per-query} & \textbf{Calls} & \textbf{Input tokens} & \textbf{Output tokens} \\
\midrule
Ours (Gemini-3-Flash)$^*$ & 14.2 & 3.2 & 91.6 & 962{,}400 & 18{,}100 \\
Ours (Qwen3-VL-235B) & 18.4 & 2.9 & 91.6 & 958{,}157 & 17{,}450 \\
\midrule
Gemini-3.1-Pro$^*$ & 0.0 & 3.5 & 39.7 & 654{,}415 & 1{,}985 \\
Qwen3.5-27B & 0.0 & 2.8 & 39.7 & 654{,}415 & 1{,}985 \\
Qwen3-VL-32B Instruct & 0.0 & 2.9 & 39.7 & 654{,}415 & 1{,}985 \\
Cosmos-Reason2-8B & 0.0 & 2.1 & 39.7 & 654{,}415 & 1{,}985 \\
\midrule
VideoMind-7B & 6.5 & 4.2 & 40.7 & 710{,}000 & 14{,}000 \\
TraveLER & 8.2 & 3.8 & 119.1 & 820{,}000 & 12{,}000 \\
SeViLA & 5.1 & 3.1 & 79.4 & 690{,}000 & 9{,}500 \\
AKS & 3.0 & 2.5 & 79.4 & 680{,}000 & 8{,}000 \\
\bottomrule
\end{tabular}%
}
\end{table}

\subsection{What counts as an object, and what a state contains}

Because the abstraction above is deliberately generic, we make three commitments explicit. \emph{(i) Entity selection.} $\mathcal{O}$ is not the set of all visible objects. An entity is registered only if it is \emph{re-identifiable} (a human annotator can point to the same entity at two separated timestamps using only visual evidence) and \emph{dynamic} (it participates in at least one state change or interaction over the video). Static background elements are excluded, which keeps $|\mathcal{O}|$ tractable and keeps the annotation burden on the entities that carry temporal content. \emph{(ii) State schema.} A state is a symbolic tuple, not a coordinate. Each $s_t^{(o)}$ is drawn from a three-part schema, \texttt{attribute} (appearance and persistent properties), \texttt{action} (the intransitive activity the entity is engaged in), and \texttt{relation} (a typed, directed link to another registered entity). We deliberately do not annotate bounding boxes: the failure mode we target is symbolic binding drift, not localisation error, and requiring boxes would restrict the benchmark to entities a detector can localise. \emph{(iii) Uncertainty and partial observability.} $\mathcal{T}_o$ is the set of timestamps at which $o$ is \emph{observed}, and is not required to be contiguous. When an entity is occluded, leaves the frame, and re-enters, annotators record the gap explicitly and then assert an identity link across it only when re-identification is supported by visual evidence (e.g., a jersey number). Otherwise the two segments remain distinct entities. Facts that an annotator can infer but not observe are never recorded, which is what makes ``the model should have known this'' a well-posed claim.

\begin{figure*}[t]
\centering
\includegraphics[width=\textwidth]{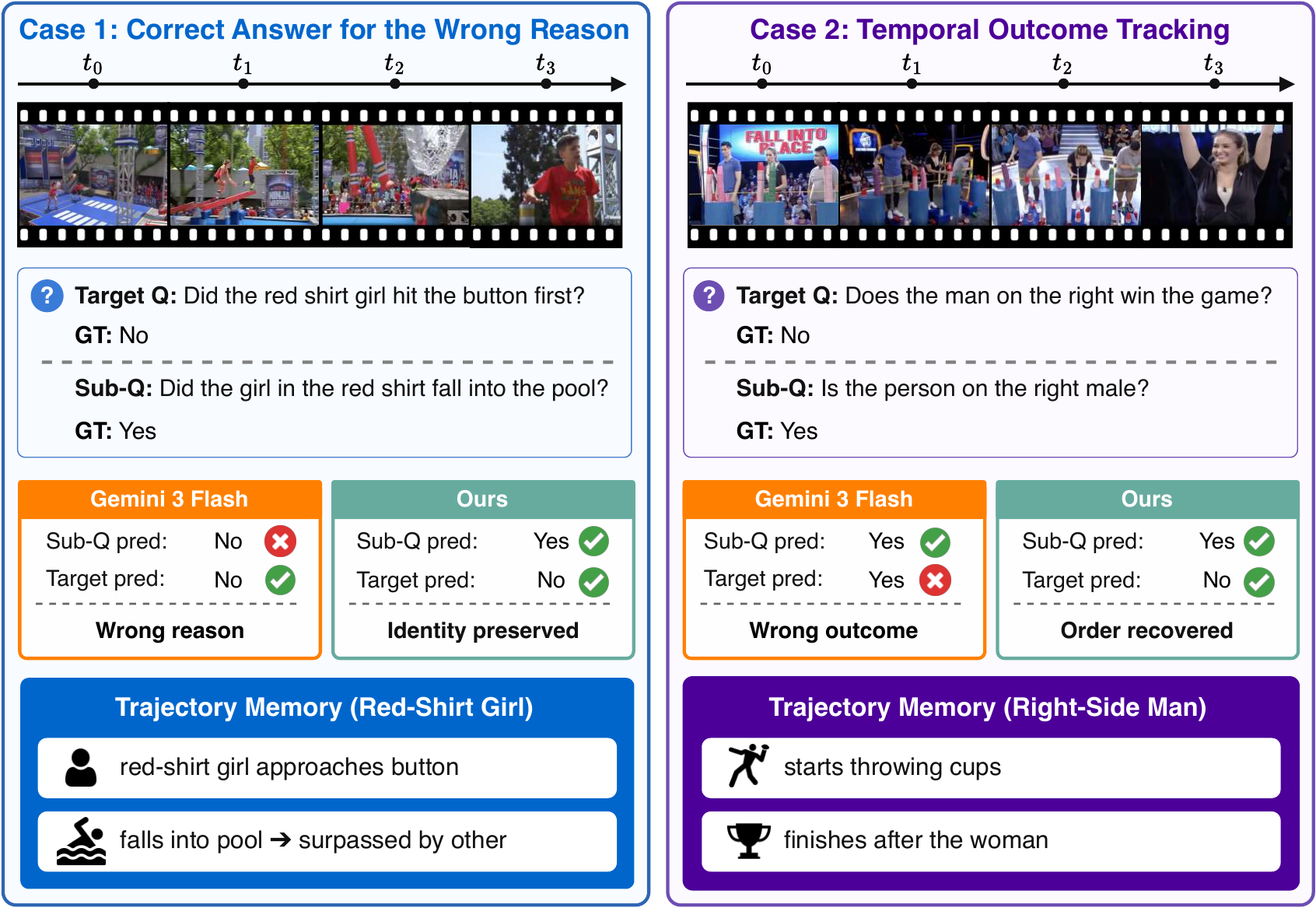}
\caption{Qualitative analysis of target and sub-question behavior.}
\label{fig:qualitative_analysis}
\end{figure*}

\subsection{Budget-aware frame-selection baselines}

The three budget-aware pipelines evaluated with a Qwen3-VL-8B backbone reach the following STRAND scores, in the order $A_{\text{faith}}$, $A_{\text{target}}$, $A_{\text{sub}}$, $A_{\text{cons}}$. AKS reaches $18.9$, $27.6$, $53.9$ and $75.6$. EFS reaches $15.9$, $24.5$, $50.2$ and $73.1$. CLIP-Retrieval reaches $16.9$, $26.2$, $52.9$ and $72.8$. All three fall below the majority-class $A_{\text{faith}}$ of $19.1\%$.

\subsection{Backbone dependence of the mechanism}

\noindent\textbf{The mechanism is backbone-dependent, and we do not overclaim it.} Read column-wise, the Gemini and Qwen ablations tell different stories, and the difference is the most informative result in this table. For Qwen3-VL-235B every component ablated in Table~\ref{tab:ablations} is load-bearing: dropping the answerer to 8 frames costs $10.4$ points, removing retrieval filtering costs $9.4$, and LLM-summarize aggregation costs $11.3$. For Gemini-3-Flash the same three interventions cost $0.6$, $0.1$, and $0.4$ points respectively, all within the confidence interval of a tie. The natural alternative hypothesis is that for a sufficiently strong backbone the benefit comes from performing \emph{structured preprocessing with a capable model at all}, rather than from object-centric trajectories or retrieval specifically, and the ablations above do not refute it. Two results do bear on it: text-only answering still reaches $73.6$ for Gemini without any frames at the answerer (Appendix~\ref{app:full_ablation_studies}, Cluster E), which shows the extracted representation carries most of the information a question needs rather than merely reorganising the answerer's visual input. Shuffling the temporal order of that same representation costs $11.8$ points ($77.1 \rightarrow 65.3$, Appendix~\ref{app:full_ablation_studies}, Cluster H), which shows the ordering of the structure, not just its existence, is what the answerer consumes. The shuffle ablation separates temporal order from unordered structured preprocessing, and it is the evidence for the narrower claim that the ordering of the representation is what the answerer consumes. It does not separate object-centric structure from generic structure, which is what the missing control below would do. We accordingly restrict the strong form of the claim, that each pipeline stage contributes, to the Qwen setting, and state the Gemini claim as the weaker one that temporal ordering of the representation is necessary.

\section{STRAND Samples}
\label{app:benchmark_sample}

\begin{figure*}[t]
    \centering
    \includegraphics[width=0.7\linewidth]{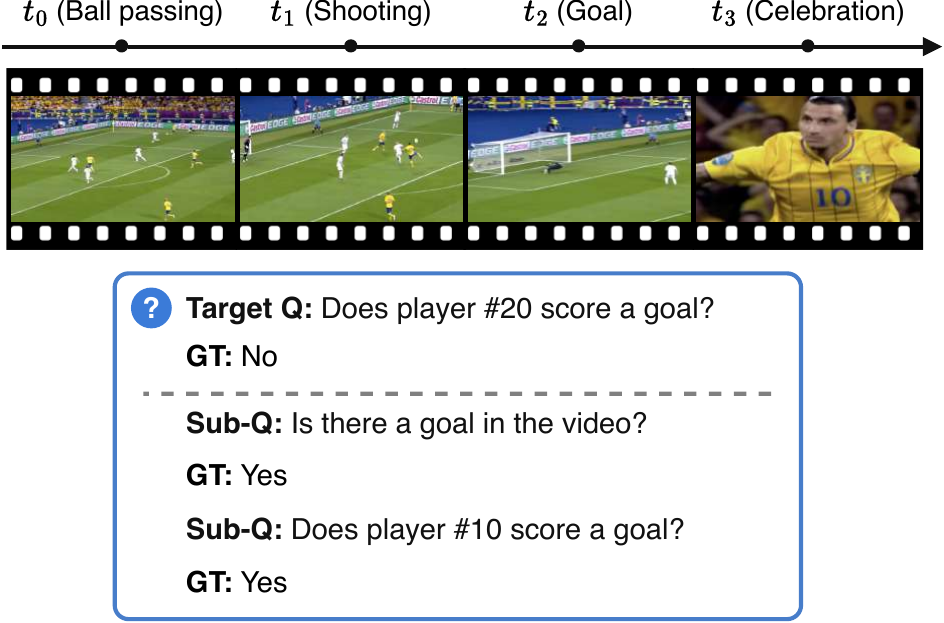}
    \caption{\textbf{Example of the STRAND dataset.} To rigorously evaluate model faithfulness, STRAND requires models to correctly answer all underlying supporting sub-questions in addition to the complex target question. This ensures the model accurately tracks temporal action sequences, from passing ($t_0$) to scoring ($t_2$), and grounds specific visual attributes (e.g., ``yellow shirt'', ``number 10'') rather than merely guessing the final answer.}
    \label{fig:app_benchmark_sample_1}
\end{figure*}

\begin{figure*}[t]
    \centering
    \includegraphics[width=0.7\linewidth]{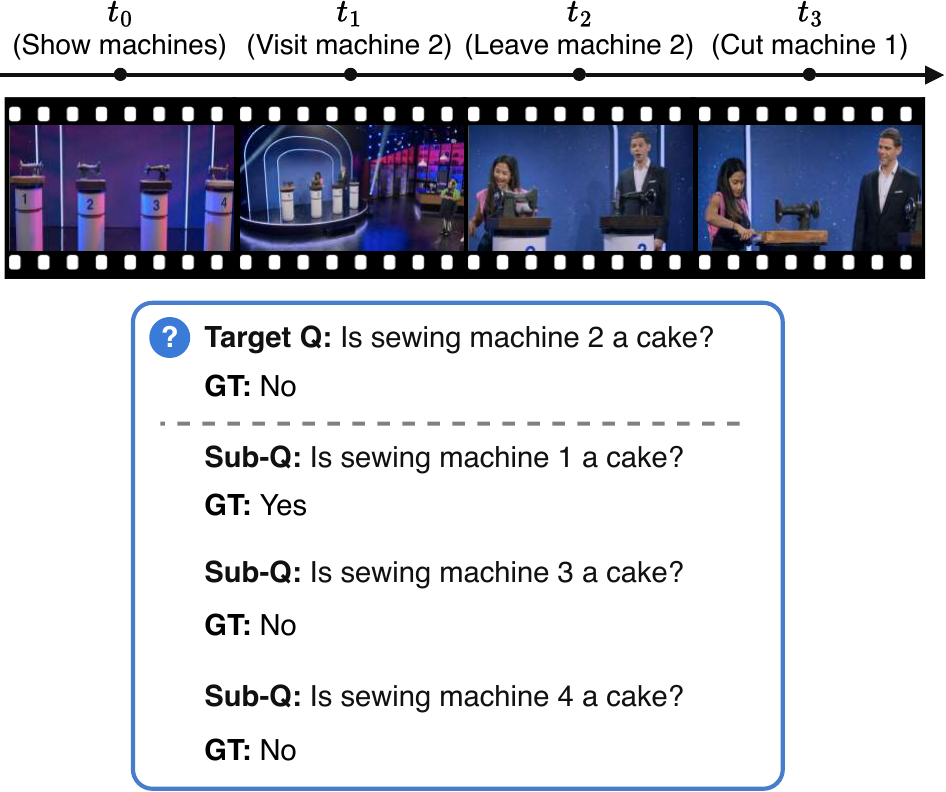}
    \caption{\textbf{STRAND evaluation on multi-step object interactions.} Building on our faithfulness criteria, models must resolve all supporting sub-questions to demonstrate true comprehension. In this scenario, the model must accurately follow a complex sequence, from the initial display of all artifacts ($t_0$) to cutting machine 1 ($t_3$), while distinguishing specific visual attributes (``number 1'', ``number 2'', ``cake'') to prevent shortcut learning.}
    \label{fig:app_benchmark_sample_2}
\end{figure*}

\begin{figure*}[t]
    \centering
    \includegraphics[width=0.7\linewidth]{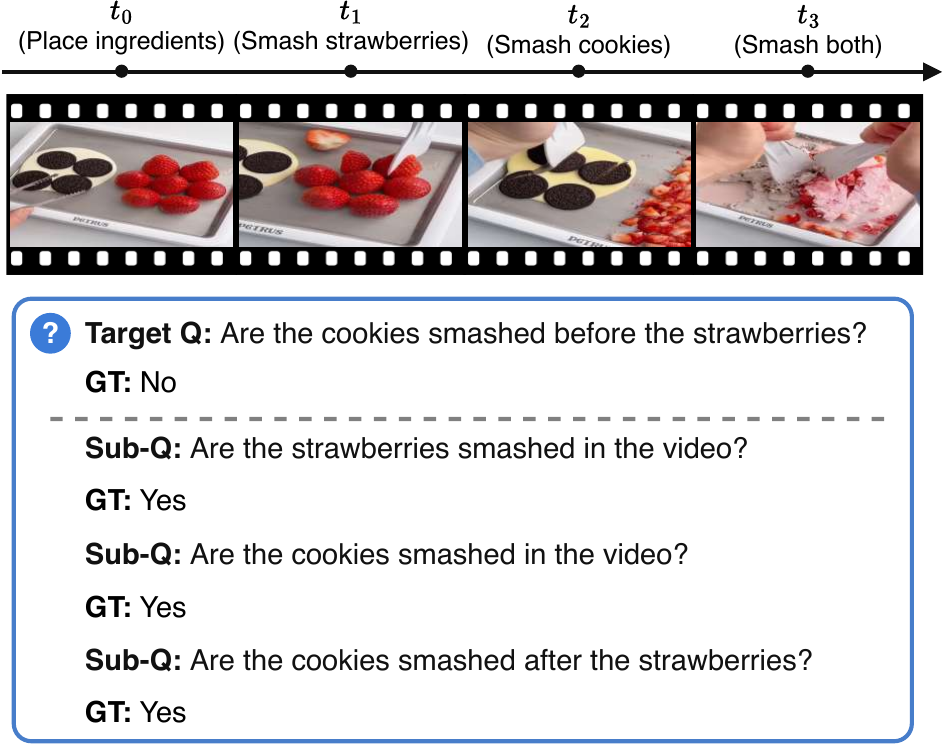}
    \caption{\textbf{STRAND evaluation of temporal ordering.} This example further illustrates how mandatory sub-questions enforce reasoning faithfulness. To succeed, the model must track the precise sequence of events, from placing ingredients ($t_0$) to smashing both items ($t_3$), and ground specific visual objects (``strawberries'', ``cookies''). This verifies that the model understands the correct temporal ordering of interactions rather than relying on educated guesses.}
    \label{fig:app_benchmark_sample_3}
\end{figure*}
\begin{figure*}[t]
    \centering
    \includegraphics[width=0.7\linewidth]{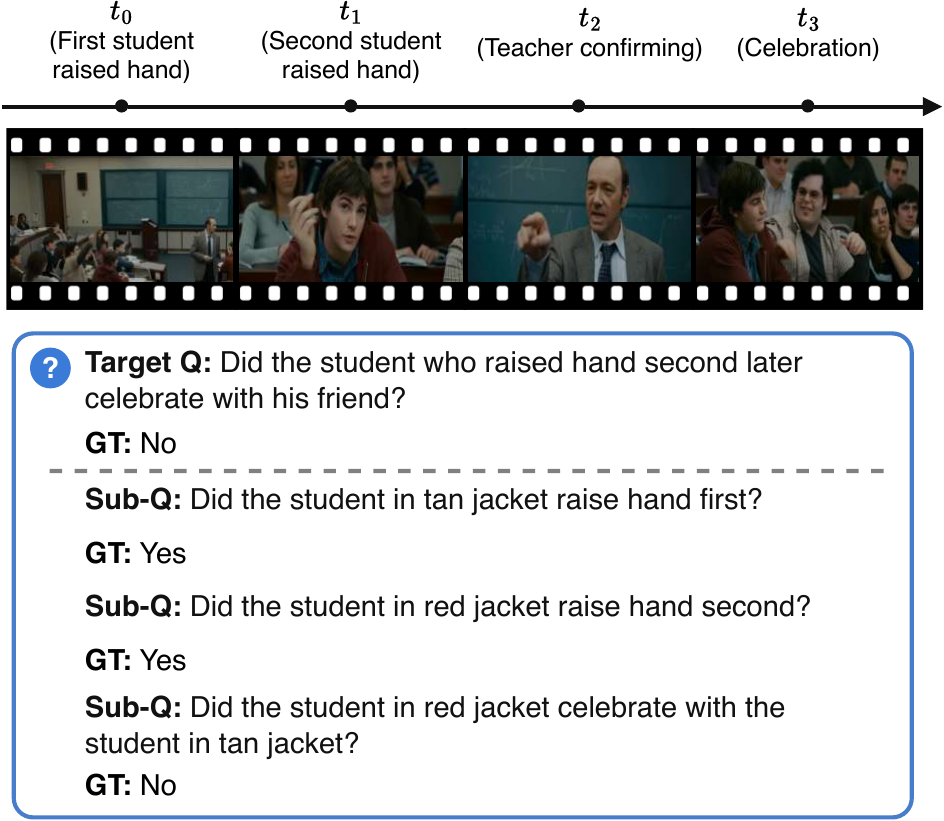}
    \caption{\textbf{STRAND evaluation of actor disambiguation over time.} This scenario highlights how STRAND tests complex temporal tracking across multiple subjects. To accurately answer the target question, the model must resolve the chronological sequence of events, from the first student raising his hand ($t_0$) and the second student raising his hand ($t_1$) to the eventual celebration ($t_3$). Mandatory supporting sub-questions force the model to explicitly differentiate between the two actors (e.g., verifying that the second student celebrated, while the first did not) to confirm precise spatiotemporal grounding rather than relying on spurious correlations.}
    \label{fig:app_benchmark_sample_4}
\end{figure*}

Figures \ref{fig:app_benchmark_sample_1} through \ref{fig:app_benchmark_sample_4} present detailed examples from STRAND. These samples showcase the complexity of the target questions and the corresponding step-by-step sub-questions required to verify a model's temporal and visual grounding capabilities.

\subsection{Additional open-weight baselines}

Four further open-weight end-to-end models were evaluated on STRAND and are omitted from Table~\ref{tab:strand_results} for space, in the order $A_{\text{faith}}$, $A_{\text{target}}$, $A_{\text{sub}}$, $A_{\text{cons}}$. Qwen3-VL-32B Instruct reaches $42.0$, $59.8$, $70.0$ and $78.1$. Qwen3.5-9B reaches $25.6$, $50.5$, $49.0$ and $57.9$. Qwen3-VL-8B Think reaches $42.5$, $61.3$, $70.1$ and $77.4$. Qwen3-VL-8B Instruct reaches $22.5$, $34.3$, $59.8$ and $77.4$. None changes the ordering of the table or the position of any bolded or underlined value.

\end{document}